\documentclass{article} %
\usepackage{iclr2027_conference,times}

\usepackage{amsmath,amsfonts,bm}

\def\eqref#1{equation~\ref{#1}}

\def\1{\bm{1}}

\DeclareMathAlphabet{\mathsfit}{\encodingdefault}{\sfdefault}{m}{sl}
\SetMathAlphabet{\mathsfit}{bold}{\encodingdefault}{\sfdefault}{bx}{n}

\usepackage[utf8]{inputenc}
\usepackage[T1]{fontenc}
\usepackage{hyperref}
\usepackage{url}
\usepackage{booktabs}
\usepackage{array}
\usepackage{amsfonts}
\usepackage{amsmath}
\usepackage{amssymb}
\usepackage{nicefrac}
\usepackage{microtype}
\usepackage{graphicx}
\usepackage{subcaption}
\usepackage{xcolor}
\usepackage{titletoc}
\usepackage{amsthm}
\newtheorem{theorem}{Theorem} \newtheorem{proposition}{Proposition}
\newtheorem{lemma}{Lemma} \newtheorem{corollary}{Corollary}
\theoremstyle{definition} \newtheorem{definition}{Definition} \newtheorem{assumption}{Assumption}
\theoremstyle{remark} \newtheorem{remark}{Remark}

\newcommand{\runsub}[2]{\refstepcounter{subsection}\paragraph{\thesubsection\hspace{0.6em}#1.}\label{#2}}

\newcommand{\Win}{W_{\mathrm{in}}}
\newcommand{\Wout}{W_{\mathrm{out}}}
\newcommand{\Ntrain}{N_{\mathrm{train}}}
\newcommand{\lmax}{\lambda_{\max}}

\newcommand{\projecturl}{\url{https://hhi-robot-learning.github.io/wave-website/}}

\definecolor{draftcolor}{RGB}{200,0,0}

\title{Object Count Generalization Is an Interface Problem: Warping a Frozen
Skill Instead of Retraining It}

\newcommand{\authorsLeft}{%
  Paul Mattes$^{1,2,*}$ \quad Ayman Abdeldayem$^{1,2}$ \quad Denis Shcherba$^{1,2}$ \\
  \textbf{Wojciech Samek$^{1,2,3}$ \quad Marc Toussaint$^{1,2,4}$} \\[3pt]
  \small $^1$Fraunhofer Heinrich Hertz Institute, Berlin, Germany \quad
  $^2$Technische Universit\"at Berlin, Germany \\
  \small $^3$BIFOLD -- Berlin Institute for the Foundations of Learning and Data, Germany \\
  \small $^4$Robotics Institute Germany \quad
  $^*$Corresponding author, \texttt{paul.mattes@hhi.fraunhofer.de}}

\author{\authorsLeft}

\iclrfinalcopy
\begin{document}

\maketitle

\begin{abstract}
Reinforcement learning policies for robot manipulation degrade when the number of objects $N$ in the scene grows beyond the training range. This is a covariate shift that also changes the dimension of the state. We propose WAVE (Warping Actions and Views of an Expert), which extends a pretrained single-object skill to many objects without updating its weights. WAVE has three parts. First, a selection policy picks the one object the robot works on next. Second, a warping policy reads a local view of the scene around that object, whose size does not depend on $N$. It emits a multiplier and an offset for every quantity the skill reads and every action it outputs. Third, the skill acts under this affine warp. It is shown a scene that holds a single object, while the warp steers it around the objects it is not shown and edits the actions that are sent to the robot. Both the selection and warping policies are trained with PPO on the multi-object reward and use the skill as a black box, so WAVE is task-general and skill-agnostic. It requires only that the skill was trained on the same embodiment and that every object has a goal pose. We also bound the number of attempts needed to place all $N$ objects, assuming positive expected net progress per attempt, and measure this progress for each object count. We evaluate WAVE on cube arrangement tasks with a gantry robot and with a Franka Panda arm. Trained with at most nine cubes, it generalizes zero-shot to several times its training count, on the gantry to scenes of up to thirty cubes. Published methods that learn multi-object manipulation show count generalization up to six objects. Fine-tuning the skill, learning from the same local view without it, and an entity transformer, even when distilled from the pretrained skill, all collapse within their training counts under the same training budget. Code,
videos and full results are available at \projecturl.
\end{abstract}

\section{Introduction}

Reinforcement learning (RL) learns a policy that maximizes the expected cumulative reward of
an agent interacting with an environment~\citep{sutton2018}. Policies trained this way for
robot manipulation solve the scene they were trained in and often degrade when that scene
changes~\citep{kirk2023}. This work addresses a covariate shift~\citep{shimodaira2000}, a change of the state
distribution under unchanged dynamics. In our case the number of objects in the scene
changes, so a policy trained with a few objects is deployed with many. Such a
shift also changes the dimension of the state, and appears for example when a
robot that has learned to manipulate a few plates is asked to fill a dishwasher. This work asks what can be learned around a frozen pretrained skill to enable zero-shot generalization to object counts unseen during training.

Two families of prior work address this shift. Set and relational architectures let a policy act
at any object count~\citep{zaheer2017,zambaldi2018}. Robot policies built this way learn the
manipulation behavior itself~\citep{li2020,haramati2024,lanier2019,li2022}. That is, a learned policy
issues the robot command at every control step, and no primitive, planner or model supplies
the motion. Such policies report whole-task success of at most $0.75$ at six objects~\citep{li2020} and at most $0.36$ at eight~\citep{park2025a,li2025dqc} on their respective benchmarks.
Further, the theory of size
extrapolation shows that networks which pool over every entity can fail to generalize to sets
larger than those seen in training~\citep{xu2021a,velikovi2025,haramati2024}. Works that keep a trained skill frozen and learn
a correction around it~\citep{silver2018,johannink2019,yuan2024,wagenmaker2025,
hirschowitz2026,zhang2023b,dcadand,policynda} absorb dynamics shifts, sensor changes and
task mismatch, but none edits both the input and the output interface of a frozen skill to
generalize across the object count.

\begin{figure}[t]
  \centering
  \includegraphics[width=0.92\linewidth]{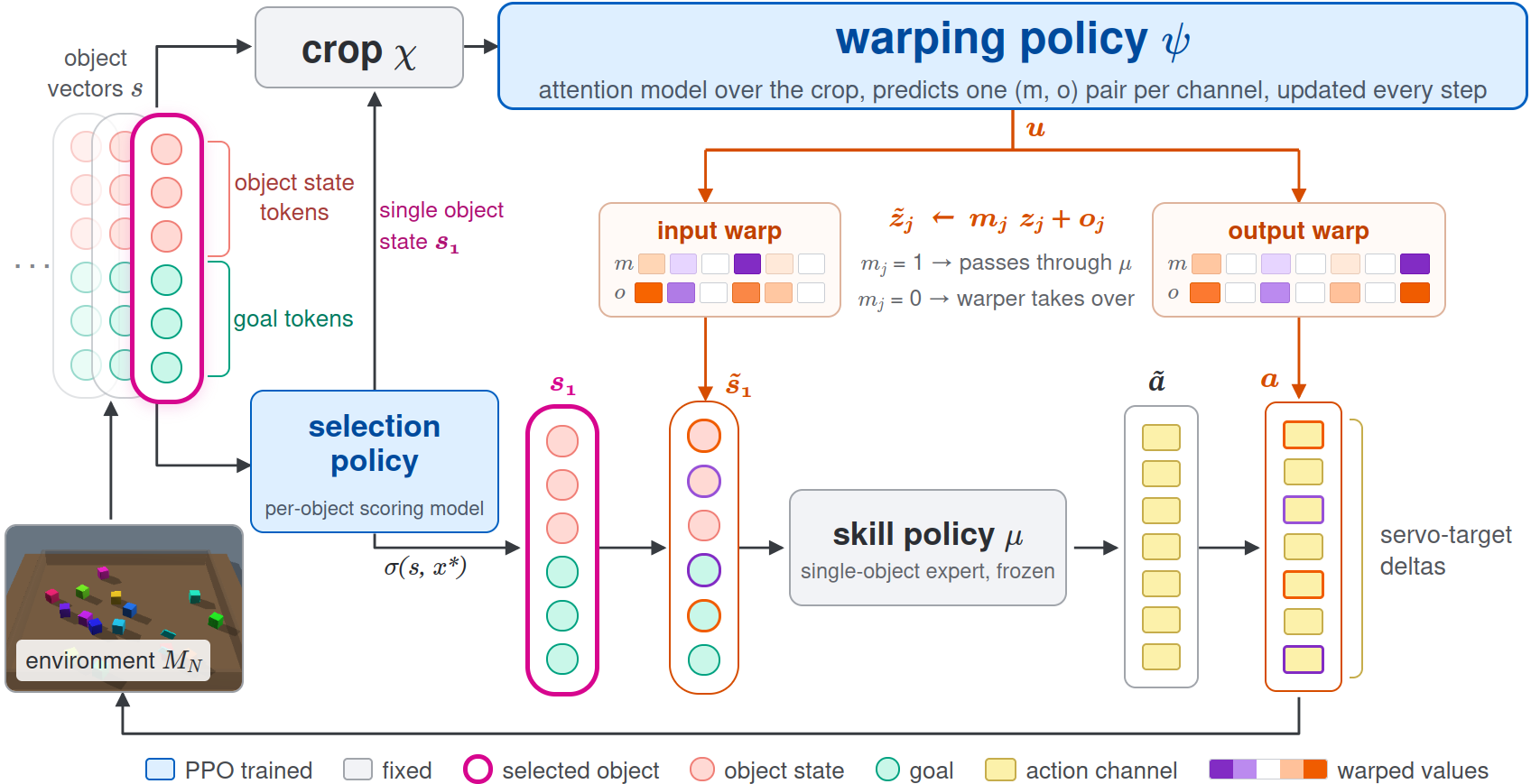}
  \caption{Overview of WAVE. The selection policy picks one object $x^*$ that is not yet placed.
  The presentation $\sigma$ shows the frozen skill $\mu$ only $x^*$ and its goal. The warping
  policy $\psi$ reads the crop $\chi$ around $x^*$ and edits every input and action channel of
  $\mu$ with a multiplier and an offset. Once $x^*$ is placed, the selection policy picks the
  next object. Only the two policies are trained.}
  \label{fig:architecture}
\end{figure}

We propose WAVE (Warping Actions and Views of an Expert), which learns a warping policy around
a frozen skill. Because the skill can already place one object, training only has to learn the
corrections needed for many.
Figure~\ref{fig:architecture} gives an overview, and our contributions are as follows.

\begin{itemize}
  \item We propose WAVE, which composes a frozen single-object skill with a learned warping
policy. The policy applies an affine edit to every input and output channel of the skill and
is trained with PPO. It uses the skill as a black box, so it applies to any goal-conditioned skill trained on the same robot and to any task that gives every object a goal pose. We show which shifts such a warp can absorb (Section~\ref{sec:absorb}).
  \item We train WAVE on six cube arrangement tasks, on a gantry robot and on a Franka Panda
arm with a gripper and with a pusher, with at most nine objects and evaluate it zero-shot
on up to thirty. On the gantry, this is two to five times the largest count to which published
methods that learn the manipulation itself generalize.
  \item We compare WAVE with ten baselines spanning skill fine-tuning, distillation, subgoal control, scratch-trained entity transformers~\citep{haramati2024}, and a fixed-per-episode input edit~\citep{zhang2023b}. They either fail beyond six cubes, even with a non-learned scripted selection of fulfillable goals, or learn their training counts but extrapolate worse. We also ablate the parts of WAVE's architecture.
\end{itemize}

\section{Related Work}
\label{sec:related}

A large body of work builds policies that are defined for any object count by making the
network permutation invariant over a set of object entities~\citep{zaheer2017,zambaldi2018}.
Entity-centric RL (ECRL)~\citep{haramati2024} learns per-object latent particles from pixels and trains an Entity Interaction Transformer with three objects that generalizes to six
objects. EC-Diffuser~\citep{ecdiffuser2025} replaces RL by behavior cloning with an entity-centric
diffusion Transformer. \citet{li2020} stack blocks with an attention-based graph network trained by a curriculum
that adds blocks, and their reported success declines beyond the counts the curriculum
visited.
Relational, curriculum and slot-based variants~\citep{mambelli2022,lanier2019,li2022,locatello2020,zadaianchuk2021} and the
successors of ECRL~\citep{ecdiffuser2025,haramati2026} report whole-task success at up to
six objects. \citet{li2020} show up to nine blocks as a figure without values,
\citet{haramati2024} show a twelve-object sort qualitatively, and the
stacking variant of that line is evaluated at two to four objects~\citep{haramati2026}.
FIOC-WM~\citep{feng2025fiocwm} learns an object-centric world model from pixels and generalizes over object attributes and skill compositions; its count test, three to eight objects, covers pushing and toggling but no stacking or other precise placement.
\citet{janisch2020} reach thirty blocks in a symbolic block world without
physics simulation. Slot models expose the count through their number of slots~\citep{zimmermann2023,krimmel2024}. Theoretical work on size extrapolation shows that networks which pool over every entity can fail on larger sets~\citep{yehudai2021,velikovi2025}, and that they can generalize when the local structure they see does not change with the count~\citep{yehudai2021,ouellette2024}. Large-scale RL learns stacking at a fixed small count~\citep{lampe2024,macaluso2026}. On the
offline cube benchmarks, which train one policy per count, the best published baseline completes the stacking goal in $0.23$ of episodes with two cubes and
in none with three or four~\citep{ogbench2024}. The hierarchical methods evaluated at eight
cubes reach at most $0.36$ averaged over five goals, from success on the rearrangement goals
and almost none on the two stacking goals~\citep{park2025a,li2025dqc}.

In task and motion planning, count generalization is easy, because a discrete search over a
known symbolic model scales to any number of objects and motion primitives supply the
manipulation~\citep{garrett2020,chitnis2022}. The same holds when a network selects scripted
primitives~\citep{zeng2020,lin2021}. Neither applies here, since we have no model and no
primitives, so the learned manipulation itself must generalize to more objects. Methods that
pass arguments to a frozen executor~\citep{riedmiller2023}
assume that the executor works for every argument.

Language and vision-language models can decompose a task into steps and call a skill for each step~\citep{ahn2022,liang2023}. In our setting, such a planner would replace the selection policy, while each step would still be executed by the skill. This execution is the limiting part, since with a scripted order over fulfillable goals the unchanged skill completes no episode with three cubes (Table~\ref{tab:exp-baselines}). A language-based planner could therefore be combined with the warp, which we leave for future work.

A single-object skill can also be reused without freezing it. Kickstarting distills a
teacher policy into a new student while the student learns the task reward~\citep{schmitt2018},
and jump-start RL rolls a guide policy for the first part of every episode and hands the rest
to an exploration policy, moving the switch point earlier as the exploration policy
improves~\citep{uchendu2022}. Both leave the multi-object behavior to a network trained on
the new task.

Another line of work keeps a trained policy or controller fixed and learns an edit of its interface. Most improve the policy on its own task and distribution, by an action residual~\citep{silver2018,johannink2019,yuan2024,ankile2024}, steered diffusion noise~\citep{wagenmaker2025}, or inserted latents and gated activations~\citep{yu2026,miao2026}. Others address a shift, such as changed dynamics~\citep{hirschowitz2026}, heterogeneous or missing sensors~\citep{dcadand}, a task mismatch around a shared backbone~\citep{policynda}, a new behavior mode~\citep{das2026}, or out-of-distribution scenes~\citep{liu2026}, but never a change in the number of objects. Closest to our approach, \citet{zhang2023b} learn diagonal projections of the state and the action of a frozen pivoting skill with a policy gradient to transfer it to objects of other shapes and sizes. Their state projection depends on a learned descriptor of the manipulated object and is constant over an episode, while their action projection also depends on the current state. WAVE keeps the frozen skill and the per-channel edits, adds an offset on every channel, and conditions both edits at every step on the multi-object scene observation. Appendix~\ref{sec:app-positioning} compares these methods with ours.

\section{Methodology}
\label{sec:method}

\runsub{Intuition}{sec:intuition}

WAVE takes a skill $\mu$ that can move a single object to its goal and uses it, without
changing its weights, to arrange many objects. A selection policy picks one object at a time,
and $\mu$ is shown a scene in which this object is the only one. The other objects are still
there, so a warping policy reads the neighborhood of the object and bends the inputs and
actions of $\mu$ where needed, for example to steer the robot around a stack or to let go of a cube on top of another
(Figure~\ref{fig:architecture}). Where nothing is in the way, $\mu$ acts unchanged.

Warping the skill is necessary because, with many objects, the behavior of $\mu$ itself has to change. Choosing the next object or sending $\mu$ to intermediate subgoals, as in hierarchical RL, leaves this behavior unchanged. For example, a subgoal can bring a cube above a stack, but it does not change how $\mu$ releases the cube there. A policy that adds a correction to the action of $\mu$ does change the behavior. However, $\mu$ still sees the unedited scene and moves along paths that collide with objects it is not shown, so the correction has to cancel part of $\mu$'s action at every step. The warp edits both what $\mu$ sees and what it outputs. Editing the input makes $\mu$ itself compute a corrected action, and editing the output produces behavior $\mu$ never learned, such as the release onto a stack.

WAVE generalizes to higher object counts for two reasons. First, $\mu$ can already place a
single object, so it is a strong exploration prior that makes multi-object tasks learnable and
training only learns the corrections needed with many objects. Second, nothing $\mu$ or the warping policy reads grows with the number of
objects. Thirty cubes produce the same kind of input as nine, whereas a policy that reads all
objects receives a new, out-of-distribution input with every added object.

\runsub{Problem and composition}{sec:problem}

Let $M_1 = (\mathcal{S}_1, \mathcal{A}, P_1, R_1, \mathcal{G})$ be a goal-conditioned manipulation
MDP with one robot and one movable object. A state $s_1 \in \mathcal{S}_1$ holds the joint
positions and velocities of the robot and the pose and velocity of the object, an action
$a \in \mathcal{A}$ is a command to the robot's actuators, $P_1(s_1, a)$ is the transition
dynamics given by the physics, a goal $g \in \mathcal{G}$ is a target pose for the object, and
$R_1(s_1, a, g)$ is a reward given for the step. Let $\mu : \mathcal{S}_1 \times \mathcal{G} \to \mathcal{A}$ be a
deterministic policy trained on $M_1$. Let $M_N = (\mathcal{S}_N, \mathcal{A}, P_N, R_N, \mathcal{G}^N)$
be the task in which the same robot must bring $N$ objects to the goal poses
$(g_1, \dots, g_N) \in \mathcal{G}^N$. Its state $s \in \mathcal{S}_N$ holds the robot and all $N$
objects, the action space $\mathcal{A}$ is that of $M_1$, $P_N$ is the physics of the full scene, in
which the objects collide with, rest on and occlude each other, and $R_N$ rewards the objects
that are at their goals (Section~\ref{sec:warper}). As illustrated in Figure~\ref{fig:architecture}, we compose $\mu$ into a policy $\pi_N$ for $M_N$ that acts on one selected object $x^*$ at a
time and computes
\begin{equation}
  \pi_N(s) = \Wout \bigl( \mu ( \Win ( \sigma(s, x^*) ; u ) ) ; u \bigr),
  \qquad u = \psi(\chi(s, x^*)),
  \label{eq:composite}
\end{equation}
with $s$ the state of $M_N$, $\sigma(s, x^*) \in \mathcal{S}_1 \times \mathcal{G}$ the presentation of
the selected object, $\chi(s, x^*)$ the crop, a local view of the scene with a fixed
width, $\psi$ the learned warping policy, and $u$ the warp parameters it
emits. $\Win$ and $\Wout$ are the warping functions that apply $u$ as per-channel affine edits of
$\mu$'s input and of its action. The
presentation $\sigma(s, x^*)$ is the state and goal $\mu$ would see in $M_1$ if $x^*$ were
the only object, with nothing about the other objects, expressed relative to the robot so
that the presented scene stays inside the distribution $\mu$ was trained on.  The warping policy runs at the same rate as $\mu$. At every control step, $\psi$ computes a new warp from the current crop before $\mu$ acts. The selection policy decides again whenever a cube is placed or a placed cube is knocked off its goal
(Section~\ref{sec:certsel}). The warping policy reads only the crop, which describes every other object in the scene relative to the selected object and its goal. The selection policy scores every object that is not yet at its goal, each together with its surroundings, so a change anywhere in the scene enters its next decision. An attempt is the period in which the selected object stays the same.

WAVE makes three assumptions about $\mu$. $\mu$ is a goal conditioned policy, it was trained
on the same robot, so that its output is a command for the robot at hand and the output side
needs no rewiring, and its input layout $\mathcal{S}_1 \times \mathcal{G}$ is known, so that the presentation
can be rendered into it. No weights, gradients, activations or knowledge of what $\mu$ does
are used.

\runsub{Selection policy}{sec:certsel}

The selection policy scores every object that is not at its goal. For each candidate $x_i$ it
reads the position of $x_i$ relative to the gripper and to its goal, the robot state, and the
neighborhood of $x_i$. A shared MLP maps these
features to two scalars per object, a priority $f_i$ and a feasibility $\hat{p}_i$ (the probability that $x_i$ can be successfully placed). The score is then computed as
\begin{equation}
  v_i = \beta \tanh f_i - \alpha \left( \tilde{d}_i + \nicefrac{\ell_i}{\hat{p}_i} \right) + \theta_i,
  \qquad x^* = \arg\max_i \, v_i,
  \label{eq:selector}
\end{equation}
where $\tilde{d}_i$ is the gripper to object distance, $\ell_i$ is an estimated length of one attempt at
$x_i$ (distance of $x_i$ to its goal plus a small constant),
and $\alpha$ a fixed travel price. Since $\nicefrac{1}{\hat{p}_i}$ is the expected number of
attempts $x_i$ takes, the bracket is the expected travel that placing $x_i$ costs. The priority head starts at 0 and the feasibility head starts at 1, so the untrained rule is shortest gripper-to-object-to-goal path first, from which the selection policy deviates only where another order yields a higher return. The priority is bounded by the constant $\beta$, so the learned priority can reorder candidates of similar expected travel but cannot overturn a travel difference larger than $2\beta/\alpha$. The feasibility head is trained on the realized outcomes of past
attempts, so an object that has a low feasibility (e.g., because it is blocked by a neighbor) has a high expected travel and is less likely to be selected, but it is not removed from the pool. Order constraints are learned, not
specified. For example, a cube can only be stacked once the cube below its slot is placed. The learned holding bonus $\theta_i$ is nonzero only for the current $x^*$, so that the selection policy keeps it until it is placed.

\runsub{The warping policy}{sec:warper}

\begin{figure}[t]
  \centering
  \includegraphics[width=1.0\linewidth]{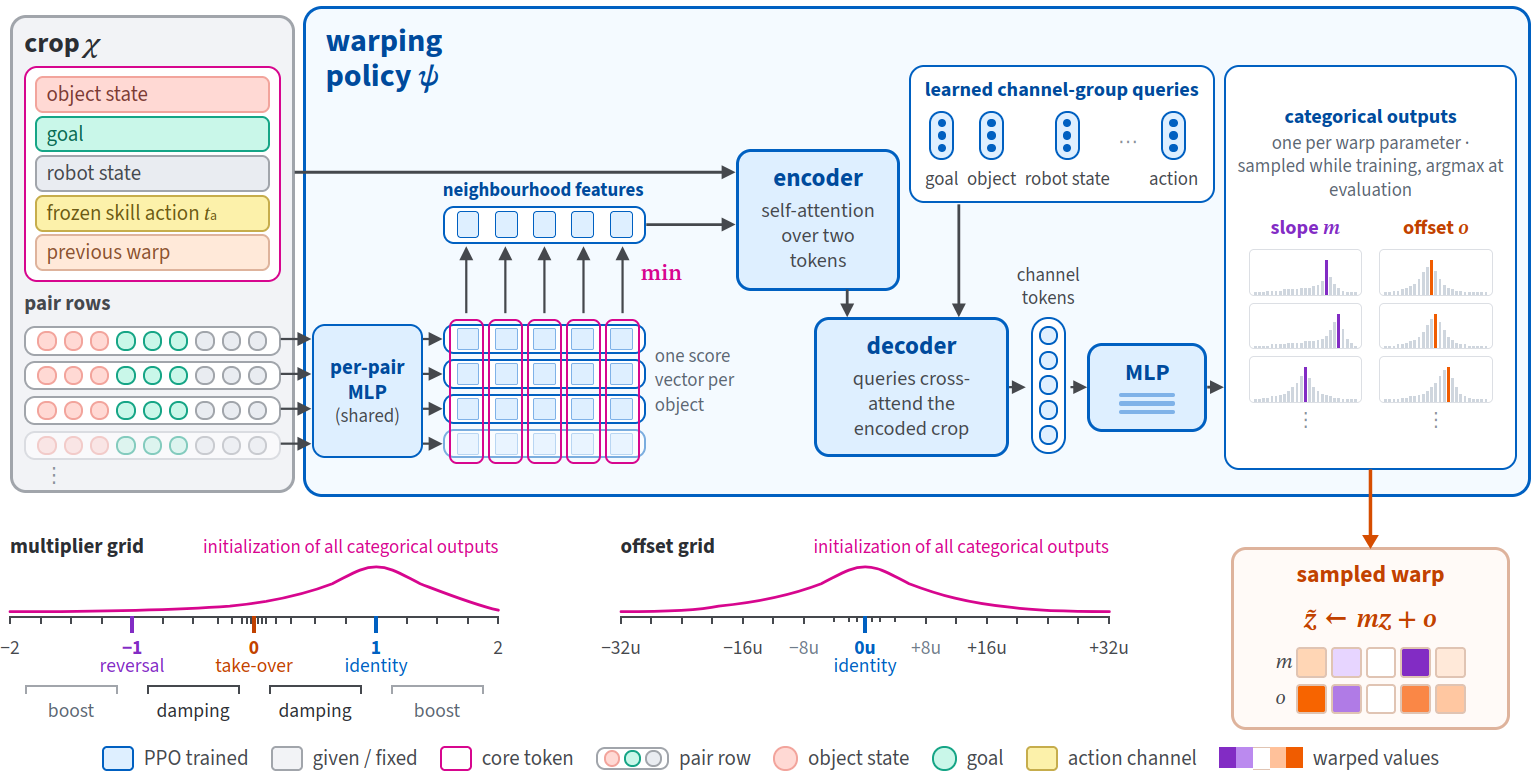}
  \caption{The warping policy $\psi$ and its initialization. The lower axis shows the multiplier
  grid, dense between take over ($m = 0$) and identity ($m = 1$), with the initial mass on the identity bin.}
  \label{fig:warper}
\end{figure}

The crop $\chi(s, x^*)$ contains the position and goal of $x^*$, the robot state and
the action of $\mu$ on the unwarped presentation $\sigma(s, x^*)$. In addition, we compute a set of geometric features for every other object that is not $x^*$. These features hold its position relative to
$x^*$, to the goal of $x^*$ and to the robot, and whether it is placed. A per-pair network maps
each object feature vector to a fixed number of bounded scores, and an elementwise minimum over these scores
reduces them to a single neighborhood encoding.
We use the minimum because the closest obstacle is the one an attempt is most likely to hit. The minimum is also independent of the count, since a neighbor five centimeters away yields the same score in a scene of three objects and in a scene of thirty. Appendix~\ref{sec:app-generalization} compares the minimum with a mean and with a learned attention over the other objects.

A channel is one scalar of $\mu$'s interface, a single coordinate of the state it reads or
of the action it emits. The channels that carry one physical quantity, the presented
goal position or the action for instance, form a channel group. The warping policy
emits the warp parameters $u$ of \eqref{eq:composite}, one pair
$(m_j, o_j)$ per channel $j$. $\Win$ and
$\Wout$ apply that pair to the channel's value $z_j$ as
$\tilde{z}_j \leftarrow m_j z_j + o_j$. At $m_j = 1$ and $o_j = 0$ the channel passes
through unchanged, at $m_j = 0$ its input is removed and $o_j$ becomes its value, so the
warping policy takes over the channel and commands it directly. As shown in Figure~\ref{fig:warper}, $\psi$ encodes the reduced pair scores and the features of $x^*$, its goal and the robot as the tokens of a small
transformer, decodes one query per channel group and emits one
categorical distribution per warp parameter over a fixed grid of values. The multiplier grid
is dense between take over and identity, the offset grid is a signed ladder in multiples
of a step size per robot (1.5\,cm on the gantry), and one further bin per channel repeats the value applied at the previous step, so the warping policy can choose to keep its previous edit. Every distribution starts with most of its mass on
the identity value, so training starts from the unwarped skill and departs from it only where
that yields a higher task reward.

The warping policy and the selection policy are trained jointly with PPO~\citep{schulman2017} on the return
of $M_N$. For each of the two policies $\pi_\phi \in \{\psi, \text{selection}\}$, PPO maximizes the clipped objective
\begin{equation}
  L(\phi) = \mathbb{E}_t \Bigl[ \min \bigl( r_t(\phi) \hat{A}_t,\;
  \mathrm{clip}\bigl(r_t(\phi), 1 - \epsilon, 1 + \epsilon\bigr) \hat{A}_t \bigr)
  + c_H \, \mathcal{H}\bigl[\pi_\phi(\cdot \mid o_t)\bigr] \Bigr],
  \qquad r_t(\phi) = \frac{\pi_\phi(a_t \mid o_t)}{\pi_{\phi_{\mathrm{old}}}(a_t \mid o_t)},
  \label{eq:ppo}
\end{equation}
where $o_t$ is the policy's input, $a_t$ its action, $\hat{A}_t$ the generalized advantage
estimate~\citep{schulman2016gae} of the return of $M_N$ from a learned value function, $\epsilon$ the clip ratio and $c_H$ an entropy coefficient. For $\psi$, $o_t$ is the crop $\chi(s_t, x^*)$, $a_t$ is the warp $u_t$ and $\pi_\phi(u_t \mid o_t)$ is the product of its per-parameter categorical probabilities. For the selection policy, $a_t$ is the selected object, drawn from a softmax over the scores $v_i$ of \eqref{eq:selector} during training, and the objective counts only the steps at which it decides. The reward $R_N$ (Appendix~\ref{sec:app-setup}) pays a small reward at every control step
proportional to the fraction of
objects currently placed, a bonus when all of them are, and a one time bonus the first time
an object is placed. The reward depends only on which objects are at their goals. It contains no term for a route, a grasp or a release and is the same for every task in Section~\ref{sec:experiments}.

\runsub{Absorbable shifts}{sec:absorb}

A warp changes the coordinates of $\mu$'s input and of its action, which makes it a
feedback transformation of the dynamics $\mu$ acts on~\citep{brunovsky1970,jakubczyk1980}. Let
$M_1'$ be the single-object task at deployment, with $\mu$'s interface unchanged and only its
dynamics $P_1'$ altered. We call the shift from $M_1$ to $M_1'$ \emph{absorbed} when the warp turns
every trajectory of $M_1'$ into one of $\mu$'s own trajectories in $M_1$,
\begin{equation}
  \Win P_1'(s_1, \Wout a) = P_1(\Win s_1, a).
  \label{eq:absorb}
\end{equation}
$\Win$ is the input warp and $\Wout$ is the output warp. An attempt under an absorbed shift then runs
like an episode of $\mu$ alone in $M_1$, only in warped coordinates. We call this a solo episode.
Its success rate is therefore $\mu$'s own, and this holds without measuring anything about $\mu$ (Appendix~\ref{sec:app-proofs}).

Which shifts a warp can absorb depends on what the warp reads. A warp that applies the same edit at every step already absorbs any recalibration of $\mu$'s sensors and actuators, since it inverts the recalibration channel by channel. On a robot whose dynamics do not change under such a shift, like the gantry, it also absorbs any translation of the scene parallel to the floor. A warp that varies with the state, as $\psi$ does through the crop, can do more. It can shift the goal that $\mu$ pursues at every step, which produces detours, hovering, retreating and holding. Such a retarget changes only the goal, so while it is held the attempt is an ordinary episode of $\mu$ toward it. The warp can also take over a channel of $\mu$ and command it directly. Take-over (multiplier 0 on an output channel) inherits no guarantee from the frozen skill. Proposition~1 in Appendix~\ref{sec:app-proofs} states which shifts such a warp can absorb. The learned warping policy is not limited to these edits and can learn any edit that raises the return, for example to compensate for small differences in the dimensions of the robot. More objects do not change what $\mu$ receives. $\sigma$ shows $\mu$ only the selected object,
and since $\sigma$ is expressed relative to the robot, a larger arena only shifts positions. What
changes with more objects is the crop, because the scene around $x^*$ becomes denser. The warping
policy reads this crop and turns the interactions with the other objects into a change of
coordinates for $\mu$.
Through these coordinate changes, the warping policy can add behaviors to $\mu$ that it never learned, such as 
curved trajectories to avoid collisions, safe release of a cube on top of another, or nudging a neighbor out of the way.
Per-channel affine maps are the minimal edit that changes single channels and contains both identity ($m=1$, $o=0$) and full takeover ($m=0$, $o$ the commanded value); since channels are physical quantities and $\psi$ acts at every step, they suffice in our experiments.
A transformation matrix in place of the per-channel edit would extend this to a robot with a
different number of joints, or to learned latent states, which we leave to future work.
Hyper-GoalNet~\citep{zhou2025hypergoalnet} instead generates the full weights of a policy from demonstrations; WAVE edits only the interface, a far smaller space that it learns from the task reward alone.

A warp cannot absorb shifts that are not observable from the crop, or shifts that it cannot invert.
A shift is \emph{unobservable} if two deployments that require different warps produce the same crop, so that a
single fixed edit has to serve both. This would occur for example when two cubes of equal size differ in friction, which the crop does not
carry. A shift is \emph{not invertible} if canceling it would require moving the state in a direction $\mu$'s
action cannot reach, so that no multiplier or offset reproduces the needed correction. This would occur for example at a kinematic singularity, with the arm stretched straight, where no joint motion moves the gripper further out whatever gain the warp applies.

\runsub{A conditional completion bound}{sec:guarantees}

Fix the object count $N$. Suppose that, conditioned on any history of the episode, the next attempt places its object with probability at least $p_N$ and knocks at most $\lambda_N$ placed objects off their goals in expectation, and never more than $\lmax$. We call $\rho_N = p_N - \lambda_N$ the net progress per attempt. If $\rho_N > 0$, then for every factor $c > 1/\rho_N$
\begin{equation}
  \Pr[\,\text{task not complete within } c N \text{ attempts}\,]
  \leq \exp\!\left( - \frac{2 (\rho_N c - 1)^2 N}{c (1 + \lmax)^2} \right)
  \label{eq:bound}
\end{equation}
(Theorem~1 in Appendix~\ref{sec:app-proofs}, from Azuma's inequality), and the expected number of attempts is at most $N / \rho_N$. The bound holds for a fixed $N$ and only under the stated assumptions, which concern every possible history of an episode, so we do not claim a time budget that is uniform in $N$. We measure $p_N$ and $\lambda_N$ per attempt as averages over evaluation episodes (Table~\ref{tab:constants}). On cube stacking, the probability $p_N$ that an attempt places the selected cube stays between $0.75$ and $0.84$ at all counts, while the number $\lambda_N$ of placed cubes it knocks off grows from $0.17$ at three cubes to $0.64$ at thirty, so the net progress $\rho_N$ falls from $+0.64$ to $+0.11$. On the pyramid (policy trained on one to ten cubes) $\rho_N$ falls from $+0.51$ to $-0.03$ and becomes negative between ten and twenty cubes, where whole-task success drops from $0.93$ to $0.19$. Since an attempt takes about 2.3\,s, the limit of 20\,s per cube allows roughly eight attempts per cube, which requires $1/\rho_N \leq 8$, a net progress $\rho_N$ of about $0.12$ or more.

\section{Experiments}
\label{sec:experiments}

\begin{figure}[t]
  \centering
  \includegraphics[width=\linewidth]{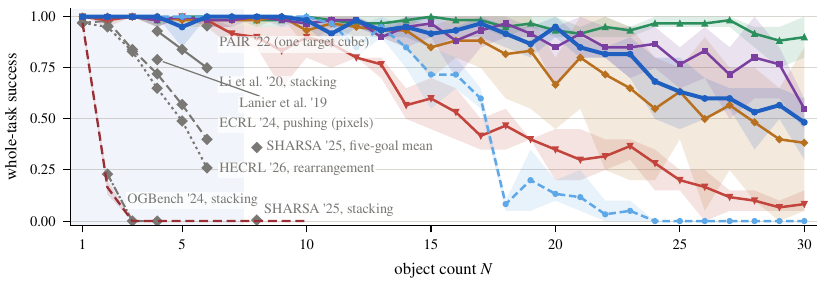}\\[1pt]
  \includegraphics[width=\linewidth]{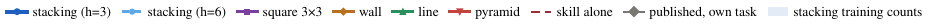}
  \caption{Whole task success against the number of cubes $N$ for WAVE on the six gantry
  tasks (color), and for published methods on their own tasks (gray, Appendix~\ref{sec:app-published}). 
  The shaded band is the training range of cube stacking
  and the dashed red curve the skill alone baseline on cube stacking ($h=3$). WAVE and skill alone are
  averaged over seeds with bands for the range. The gray series come from other tasks, simulators and embodiments and are shown for context only.}
  \label{fig:results}
\end{figure}

We evaluate WAVE on six cube arrangement tasks, stacks of height $h=3$ and $h=6$, a line, a
square, a pyramid and a wall (Figure~\ref{fig:app-snapshots}). A gantry robot with a
parallel-jaw gripper runs every task with one to thirty cubes (Section~\ref{sec:exp-count}). A
Franka Panda arm runs five of them with a parallel-jaw gripper or a sphere-tip pusher, up to fifteen, twenty or thirty cubes (Section~\ref{sec:exp-panda}). Each policy trains on mixed counts of one to
$\Ntrain$, with $\Ntrain$ between six and nine depending on the task and robot
(Appendix~\ref{sec:app-setup}). An episode ends as a success as soon as every cube is within the position tolerance of its goal and on its intended support. This holds in training and in evaluation, so the last cube can still be in the gripper. We call the fraction of successful episodes \emph{whole-task success}. We show the results for the gantry robot on stricter success criteria in~\ref{app:held_scores}.
For every object count evaluation, we use initial layouts which were not part of the training pool. Every evaluation uses a deterministic policy, 20 s per cube and three seeds. The robot, the
skill, the tasks, all hyperparameters, and the full result tables are given in
Appendices~\ref{sec:app-setup} and~\ref{sec:app-results}.

\runsub{Count generalization on the gantry}{sec:exp-count}

Trained on one to seven cubes, WAVE (Figure~\ref{fig:results}) stays at or above $0.92$ on
cube stacking ($h=3$) up to twenty cubes and falls to $0.48$ at thirty. The line, square,
pyramid and wall complete between $0.80$ (pyramid) and $0.97$ (line and square) at twice
their training count and between $0.08$ (pyramid) and $0.90$ (line) at thirty
(Table~\ref{tab:full-results}).

Published methods that learn the manipulation itself reach at most $0.75$ at six
objects~\citep{li2020} and $0.36$ at eight~\citep{park2025a,li2025dqc}, while every WAVE task is
at or above $0.98$ at six. WAVE stays above the $0.75$ of \citet{li2020} up to twenty-three cubes on cube stacking,
the task closest to theirs, a nearly fourfold extension, and up to thirty on the line, a
fivefold one. These comparisons cross tasks, simulators, embodiments and in places success
criteria (Appendix~\ref{sec:app-published}), since no controlled comparison exists at these
counts.

Cube stacking with $h=6$ tests placement precision. The training scenes hold at most seven
cubes, one full stack and the base cube of a second, so no training episode places a cube onto
the second stack. WAVE still builds the second stack up to five cubes high and completes $0.97$
or more of the episodes up to eleven cubes. Success falls to $0.85$ at twelve cubes, where the second stack needs its sixth cube, and is zero from twenty-four on, lost to toppled stacks rather than missed placements.

At high counts, episodes fail mostly by running out of time rather than by losing cubes (Table~\ref{tab:end-causes}). On cube stacking ($h=3$), on average $26.8$ of $30$ cubes are placed at the same time at the peak, but placed-to-unplaced transitions grow from $7.5$ per episode at ten cubes to $126$ at thirty. The policy re-places most of these cubes, which costs time and shrinks the net progress per attempt (Section~\ref{sec:guarantees}). At thirty cubes, $0.49$ of the episodes end at the time limit and $0.06$ because a cube left the workspace. With $60$ s instead of $20$ s per cube, success at thirty cubes rises from 0.48 to 0.67 on stacking and from 0.08 to 0.27 on the pyramid..

Appendix~\ref{sec:app-generalization} reports further generalization experiments with the unchanged skill. One pair of policies trained on five tasks at once completes all five. WAVE also handles larger cubes, boxes and scenes with objects of different sizes and shapes, and it still completes its tasks when disturbances set cubes to other positions during the episode. Success decreases beyond five cubes for cubes with 0.8 times the edge, and when a built structure of six or more cubes has to be taken apart without a cube falling.

\runsub{Count generalization on the Panda arm}{sec:exp-panda}

\begin{figure}[t]
  \centering
  \begin{minipage}[c]{0.50\linewidth}
    \centering
    \includegraphics[width=\linewidth]{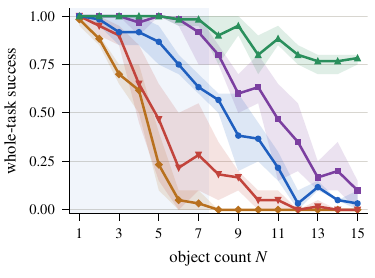}\\[1pt]
    \includegraphics[width=\linewidth]{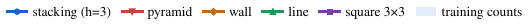}
  \end{minipage}\hfill
  \begin{minipage}[c]{0.48\linewidth}
    \centering
    \scriptsize
    \setlength{\tabcolsep}{1.5pt}
    \begin{tabular}{@{}cc@{}}
      gantry, fifteen cubes & Panda, ten cubes \\
      \includegraphics[height=0.32\linewidth]{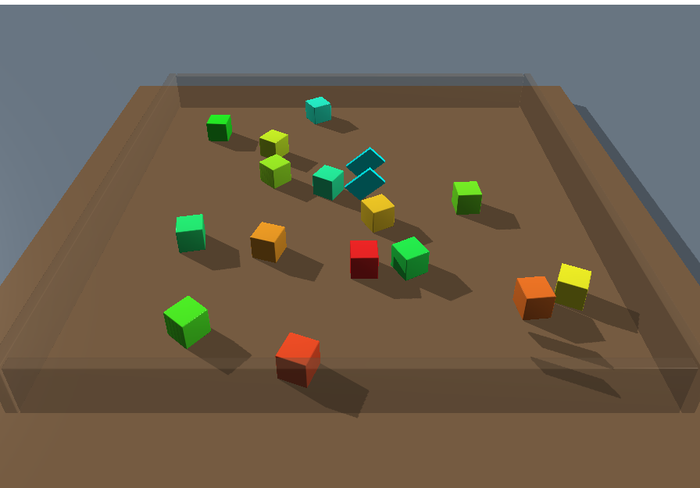} &
      \includegraphics[height=0.32\linewidth]{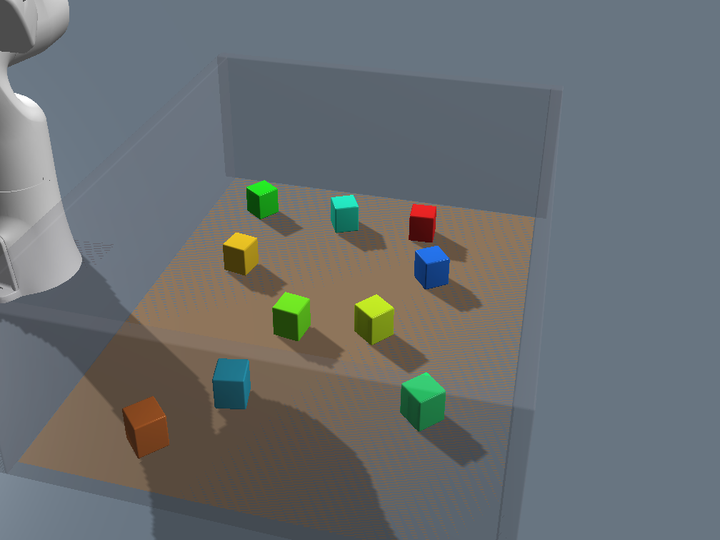} \\[1pt]
      \includegraphics[height=0.32\linewidth]{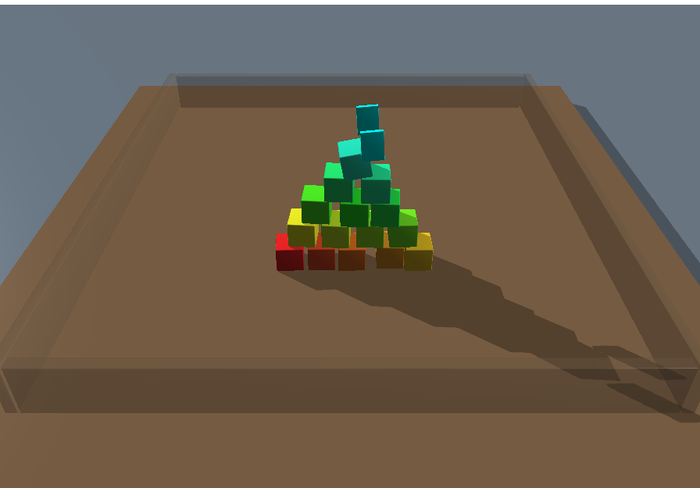} &
      \includegraphics[height=0.32\linewidth]{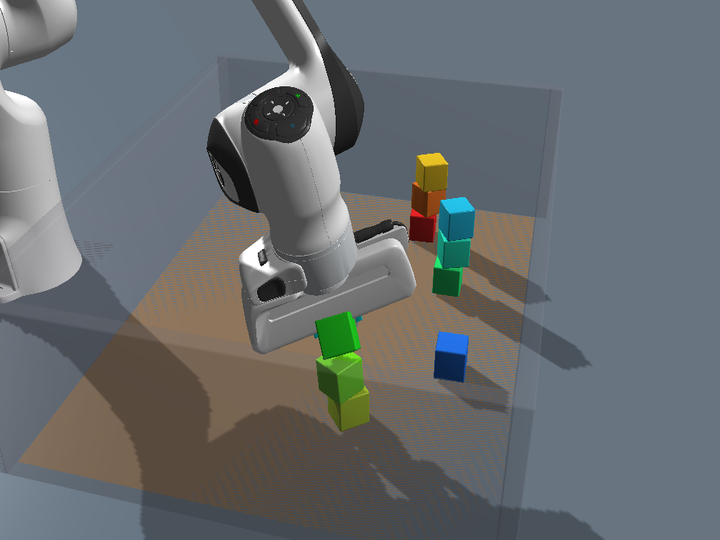} \\
      pyramid & cube stacking $h{=}3$ \\
    \end{tabular}
  \end{minipage}
  \caption{Left, whole-task success against the object count on the Panda arm, the first fifteen counts of every task. Right, a start scene (top)
and a completed episode (bottom) on each robot, a fifteen-cube pyramid on the gantry and ten-cube
stacking ($h = 3$) on the Panda. All tasks are rendered in Figures~\ref{fig:app-snapshots}
and~\ref{fig:app-panda-snapshots}.}
  \label{fig:panda}
\end{figure}

We transfer the same method and reward to a Franka Panda arm with 4\,cm cubes in two
embodiments, each with its own frozen skill (Appendix~\ref{sec:app-setup}). A parallel-jaw
gripper handles cube stacking ($h=3$), the pyramid and the wall. A sphere-tip pusher, whose
skill pushes one cube to a planar position and yaw goal, handles the line and the square.
Figure~\ref{fig:panda} shows the first fifteen counts and Table~\ref{tab:panda-results} the
full range of each task.

With the pusher, WAVE completes $0.95$ of the line episodes at nine cubes, its largest
training count, $0.78$ at fifteen and $0.62$ at twenty, and failed episodes end only a cube
or two short ($14.2$ of fifteen cubes placed on average on these episodes). The square is harder, $0.60$ at nine
and $0.10$ at fifteen, since its interior slots leave the pusher no free approach side and
late cubes are pushed in across placed ones until the episode runs out of time.

With the gripper, success already falls within the training range. Cube stacking completes
$0.75$ at six, level with the best published result~\citep{li2020}, and $0.37$ at ten, while
the pyramid and the wall complete $0.22$ and $0.05$ at six. Both failure modes of
Section~\ref{sec:exp-count} appear early. At six cubes the wall ends $0.76$ of its episodes by
timeout, and on cube stacking the fraction of placed cubes that stay placed falls from $0.95$
at six cubes to $0.66$ at fifteen. This is likely due to the large body of the arm, which knocks neighbors off their goals when cubes are placed next to each other. Longer
training, larger networks or a richer crop may close this gap and are left for future work.

We also deployed a cube stacking policy trained in simulation on a physical Panda arm, with cube poses tracked from AprilTag fiducials by four calibrated RGB cameras. It placed all five cubes of a two-tower task in 8 of 10 runs, and a policy fine-tuned for five-cube towers built a five-high tower (Appendix~\ref{sec:app-realworld}).

\runsub{Baselines}{sec:exp-baselines}

\begin{table}[t]
  \caption{Whole task success of WAVE and seven of the ten baselines on the gantry, the mean over three seeds. The first three columns give how each method uses $\mu$, which objects besides $x^*$ it sees, and how it picks the next object. Every seed of every other baseline in this table is at $0.00$ at six and at thirty cubes. Bold marks the highest mean per count. Seed ranges, every count and the other three baselines are in Table~\ref{tab:baselines}.}
  \label{tab:exp-baselines}
  \begin{center}
  \fontsize{7.5}{9}\selectfont
  \setlength{\tabcolsep}{2.4pt}
  \begin{tabular}{@{}llll@{\hspace{7pt}}ccccc@{\hspace{7pt}}ccccc@{}}
    \toprule
    & & & & \multicolumn{5}{c}{line, $N$} & \multicolumn{5}{c}{cube stacking ($h=3$), $N$} \\
    \cmidrule(lr){5-9} \cmidrule(l){10-14}
    & $\mu$ & other objects & selection & 1 & 2 & 3 & 6 & 30 & 1 & 2 & 3 & 6 & 30 \\
    \midrule
    WAVE & frozen & crop & learned & \textbf{1.00} & \textbf{1.00} & \textbf{1.00} & \textbf{1.00} & \textbf{0.90} & \textbf{1.00} & \textbf{1.00} & \textbf{1.00} & \textbf{1.00} & \textbf{0.48} \\
    \midrule
    input edit fixed per episode & frozen & crop & learned & \textbf{1.00} & 0.93 & 0.98 & 0.75 & 0.25 & \textbf{1.00} & 0.98 & \textbf{1.00} & 0.90 & 0.00 \\
    skill alone & frozen & none & scripted & 0.79 & 0.35 & 0.00 & 0.00 & 0.00 & 0.97 & 0.17 & 0.00 & 0.00 & 0.00 \\
    TD7 on the crop & -- & crop & scripted & 0.00 & 0.00 & 0.00 & 0.00 & 0.00 & 0.00 & 0.00 & 0.00 & 0.00 & 0.00 \\
    fine-tuning $\mu$ on the crop, $\omega = 1$ & fine-tuned & crop & scripted & 0.96 & 0.43 & 0.03 & 0.00 & 0.00 & 0.98 & 0.64 & 0.03 & 0.00 & 0.00 \\
    HRL subgoals & frozen & crop & learned & 0.54 & 0.18 & 0.00 & 0.00 & 0.00 & 0.83 & 0.53 & 0.10 & 0.00 & 0.00 \\
    EIT~\citep{haramati2024} & -- & all & -- & 0.20 & 0.00 & 0.00 & 0.00 & 0.00 & 0.53 & 0.00 & 0.00 & 0.00 & 0.00 \\
    EIT distilled from $\mu$ & distilled & all & -- & 0.82 & 0.56 & 0.37 & 0.00 & 0.00 & 0.94 & 0.20 & 0.00 & 0.00 & 0.00 \\
    \bottomrule
  \end{tabular}
  \end{center}
\end{table}

We compare WAVE on the line and on cube stacking ($h=3$) with ten baselines, trained on the same gantry, task, reward and training counts, and with the same $\mu$ wherever one is used. Table~\ref{tab:exp-baselines} reports seven of them and Appendix~\ref{sec:app-baselines} all ten. Apart from HRL subgoals and the fixed input edit, which keep WAVE's selection policy, the baselines that act on one object at a time use a scripted nearest-first rule that offers only objects whose goal is currently fulfillable, an easier task than the one WAVE learns from the reward. Nine of the ten baselines fail already within their training counts and complete at most $0.02$ of the episodes with six cubes and none with more. For these, the comparison shows that they do not learn the multi-object task with our training budget, and it cannot tell how they would extrapolate if they did. Only the fixed input edit learns its training counts and thereby tests extrapolation.

TD7 on WAVE's crop without $\mu$ completes no episode at any count, and the Entity Interaction
Transformer (EIT) of \citet{haramati2024} learns only the one-cube task, $0.53$ on cube
stacking and $0.20$ on the line. Distilling $\mu$ into the EIT raises its success on the line to $0.82$ at one cube and $0.37$ at three, but it still fails within its training counts. We did not isolate the cause of either failure.

Fine-tuning $\mu$ on its own input and JSRL~\citep{uchendu2022} retain most of the one-cube skill ($0.72$ and $0.45$ on the line, $0.88$ and $0.87$ on cube stacking) and improve on the skill alone at two cubes on cube stacking ($0.42$ and $0.45$ against $0.17$), but complete at most $0.13$ of the episodes with three or more cubes. Fine-tuning on the crop without an anchor completes no episode at any count. Anchoring the actor to $\mu$~\citep{fujimoto2021} with $\omega = 1$ keeps the skill and improves on it at two cubes ($0.64$ against $0.17$ on cube stacking) but drops to $0.03$ at three. The anchor penalizes every deviation from $\mu$ equally, while a warp can remain at the identity on some channels and steps and take over others ($m_j = 0$).

HRL subgoals, which moves only the goal $\mu$ pursues, completes $0.53$ of the two-cube stacking episodes and at most $0.10$ from three cubes on. A subgoal changes where $\mu$ goes but not how it releases a cube onto a stack, the phase that needs the output side of the warp (Appendix~\ref{sec:app-necessity}). Warping only $\mu$'s input channels, which
also edits what $\mu$ sees of the robot and of the object, nearly matches the full warp,
$0.81$ against $0.82$ mean success (Section~\ref{sec:exp-Ablation}).

As a further baseline we apply the interface of \citet{zhang2023b}, whose state projection is constant over an episode, to our setting. We train WAVE with the input edit chosen at the first step of an episode and kept until its end and change nothing else, so the output edit is still chosen at every step, and the offsets, the crop and the learned selection policy, which are not part of their method, remain. This baseline learns its training counts (at least $0.83$ on cube stacking and $0.75$ on the line) but extrapolates worse than WAVE. On cube stacking it completes $0.60$ of the episodes at ten cubes and $0.08$ at twenty, compared to $0.98$ and $0.95$ for WAVE (Figure~\ref{fig:snapshots}). On the line it completes $0.55$ and $0.35$, compared to $1.00$ and $0.93$.

\runsub{Ablations}{sec:exp-Ablation}

We ablate WAVE on cube stacking ($h=3$), the square and the pyramid. First, we replace the
selection rule of \eqref{eq:selector} with a random choice or a nearest-first choice over the objects whose goal is currently fulfillable, a filter the learned selection policy does not use. Second, we let the warp edit only the input or only the output channels of $\mu$.
Editing only the input is close to methods that remap observations~\citep{dcadand,liu2026}, and
editing only the output is close to residual and action distribution
methods~\citep{silver2018,johannink2019,hirschowitz2026}. Third, we let the warp emit only
offsets or only multipliers. Random selection completes no episode with five or more cubes on cube stacking and the pyramid, or with nine or more on the square, while the
fixed nearest-first rule matches the learned selection policy on stacking and the square and
falls behind it on the pyramid. Warping only the
input channels nearly matches the full warp, warping only the output channels collapses by
ten cubes, and warps with only multipliers or only offsets fall in between, multipliers above offsets.
Curves, analysis and the constants of \eqref{eq:bound} are in
Appendix~\ref{sec:app-ablations}.

To measure when in an episode each side of the warp is needed, we stop a running episode at every time step, copy the simulator, and
continue the copies with either the input or the output warp switched off. We then measure how
much longer the current phase of the manipulation (approach, grasp, carry, release or depart)
takes compared to copies where the warp stays on. Switching off the output warp delays mainly the release, since the frozen skill never releases a cube on its own. Switching off the input warp
slows down every phase, by $0.95$--$1.56$\,s, because it steers the skill throughout the
episode. Appendix~\ref{sec:app-necessity} gives the details and shows an example episode.

\section{Conclusion and Limitations}
\label{sec:conclusion}

We introduce WAVE, which extends a frozen single-object skill to multi-object arrangement tasks. It shows the skill
one object at a time, learns a warping policy that edits every channel of the skill's
interface, and learns a selection policy that finds from outcomes which object to attempt next. The skill is a black box, no learned component observes the object count, and
all six tasks use the same reward and hyperparameters. Trained with at most nine cubes, WAVE
generalizes on the gantry to two to five times the largest count reached by published methods
that learn the manipulation itself, whereas fine-tuning the skill, a policy on the same local
view and an entity transformer, even when distilled from the skill, collapse within their
training counts under the same training budget. In the main experiments, one warping policy and one selection policy are trained per task.

We evaluated WAVE in simulation on three embodiments and demonstrated the real-world transfer
of a multi-object stacking policy on a physical Panda arm (Appendix~\ref{sec:app-realworld}). The Panda gripper tasks degrade within their training
range (Section~\ref{sec:exp-panda}). The method applies only to tasks that
provide a goal pose for every object. The main experiments use identical cubes, and an object class that the skill cannot manipulate alone would need its own skill. The policies do not anticipate objects that move on their own. The skill moves one object at a time, so tasks that need the coordinated motion of several
objects, or in which objects affect each other without touching, are out of scope.

\subsection*{Acknowledgements}

This work has received support from the German Federal Ministry of Research, Technology and
Space (BMFTR) under the Robotics Institute Germany (RIG) and as grant BIFOLD (01IS18025A,
01IS180371I).

\subsection*{AI use statement}

In this work, we used generative AI tools for related work surveys, implementations, conduction of experiments and drafting parts of this paper. We have not used generative AI tools for
summarization and reading of related papers, the ideation of our approach, the generation of plots and for the final version of this paper. We have reviewed all AI-assisted work.
We take responsibility for the final content of this work, including text, claims or
artifacts produced with the aid of generative AI.

\subsection*{Ethics statement}

WAVE is basic research on making manipulation policies generalize across object counts, and
every result in this paper except one demonstration on a physical Panda arm is simulated. The risks it touches are the generic ones of more
capable robot manipulation, displaced manual handling labor and the possibility of building
manipulation into systems used for harmful ends, and none is specific to this method or
amplified by it. On the positive side, reusing a trained skill instead of retraining from
scratch lowers the compute cost of extending a robot's competence.

\subsection*{Reproducibility statement}

Appendix~\ref{sec:app-setup} specifies the robot, the simulation, the frozen skill and its
training, the six tasks and their goal predicate, the reward with all its constants, every
hyperparameter of the warping and selection policies and the evaluation protocol.
Appendix~\ref{sec:app-results} reports the full per count results with seed ranges,
Appendix~\ref{sec:app-ablations} describes how the constants of \eqref{eq:bound} are measured,
and Appendix~\ref{sec:app-proofs} states every assumption of Proposition~1 and Theorem~1 and
gives the complete proof. Code, videos and full results are available at \projecturl, and the
supplement contains the implementation and the code that reproduces our results.

\bibliography{main}
\bibliographystyle{iclr2027_conference}

\appendix
\clearpage
{\noindent\huge\bfseries Appendix\par}
\vspace{1.5em}
{\noindent\Large\bfseries Table of Contents\par}
\vspace{0.3em}\hrule\vspace{0.8em}
\newcommand{\appendixentry}[3]{%
  \noindent\textbf{\ref{#1}\hspace{0.8em}#2}\dotfill\pageref{#1}\par
  \vspace{0.15em}%
  \noindent\hspace{1.7em}\parbox[t]{\dimexpr\linewidth-1.7em\relax}{\small #3}\par
  \vspace{0.7em}}
\appendixentry{sec:app-setup}{Experimental setup}{The robots and the simulation, the frozen
  skill, the six tasks with the rule for when a cube counts as placed, the training
  hyperparameters, the reward and the evaluation protocol.}
\appendixentry{sec:app-results}{Full results}{Whole-task success at every object count for all
  gantry and Panda tasks, the curves and the table of all ten baselines, and how episodes end
  (success, time limit or a cube that left the workspace).}
\appendixentry{sec:app-realworld}{Real-world Panda experiments}{A cube stacking policy trained
  in simulation and run on a physical Panda arm with camera-based pose tracking, with ten
  consecutive rollouts and a tower of five cubes.}
\appendixentry{sec:app-baselines}{Baselines}{How each of the ten baselines is built and
  trained.}
\appendixentry{sec:app-published}{Overview of published methods and results}{The published
  multi-object results that Figure~\ref{fig:results} shows in gray, with their tasks,
  embodiments and success criteria.}
\appendixentry{sec:app-ablations}{Ablations and measured constants}{Ablations of the selection
  rule, of the two sides of the warp and of its multipliers and offsets, and the measured
  quantities of the completion bound per attempt.}
\appendixentry{sec:app-necessity}{Interventional estimation of warp necessity}{The input or the
  output warp is switched off in copies of a running episode to measure in which phase of the
  manipulation each side is needed.}
\appendixentry{sec:app-proofs}{Formal statements and proofs}{The assumptions, Proposition~1 on
  when an attempt of the warped skill is an episode of the skill alone, and Theorem~1 on the
  number of attempts that complete a task, with proofs.}
\appendixentry{sec:app-positioning}{Positioning of interface-editing methods}{A comparison of
  WAVE with prior methods that edit the input or the output of a frozen policy.}
\appendixentry{sec:app-generalization}{Generalization experiments}{Eight further experiments:
  one pair of policies for five tasks, other object sizes and shapes, cubes that have to stay
  in place, disturbances, episodes that start from a built structure, stricter success
  criteria, and replacements of the minimum over the other objects.}
\vspace{0.2em}\hrule\vspace{1.5em}

\section{Experimental setup}
\label{sec:app-setup}

\emph{Robot and simulation.} The robot is a three axis gantry with a three joint wrist
and a parallel-jaw gripper (8 joints, 7 actuators). It is commanded at 20 Hz with a 7-D
action (three translations, three wrist rotations, one grip command). The scene is simulated
in MuJoCo~\citep{todorov2012} on the GPU with 2 ms physics substeps and 2048 parallel
environments. Objects are cubes of 16 cm edge length and 0.2 kg in a
2.7 m $\times$ 2.7 m workspace. The Panda experiments of Section~\ref{sec:exp-panda} use a
simulated 7-DoF Franka Panda arm commanded by joint velocities, with cubes of 4 cm edge
length, in two embodiments, one with a parallel-jaw gripper and a binary grip channel (8-D
action) and one with a rigid sphere-tip in place of the jaws (7-D action) that pushes cubes
across the table and never lifts them.

\emph{The skill $\mu$.} $\mu$ moves one cube from any resting pose to any goal pose,
including the goal orientation. We train it with TD7~\citep{fujimoto2023} in a single-object scene with constrained sampling and reinforcement learning
(CSRL)~\citep{toussaint2026combined}. Its input is the 65-D
state vector of joint and cube positions, joint and cube velocities, the servo errors of the 7
actuators, the positions of four cube corners, the goal-relative rotation and the corner to
goal offsets, and its action is the 7-D command above. $\mu$ is frozen in every experiment
except the fine-tuning baseline. On its own task it succeeds in $0.97$ of 64 evaluation
episodes at the checkpoint we freeze. Each Panda embodiment has its own skill, trained the
same way in its own single-object scene. The gripper skill places one cube as the gantry
skill does, and the pusher skill brings one cube to a position and yaw goal in the plane,
where the cube's fourfold symmetry identifies yaws a quarter turn apart.

\emph{Tasks.} The six tasks share one goal predicate and differ in the goal layout.
The pyramid, square, line and the wall grow with $N$, and the stacks are repeated at their target height. Every task is trained and evaluated in the 2.7 m $\times$ 2.7 m workspace, except that the line is evaluated in a workspace widened to fit thirty cubes. A cube counts as placed when it is within the tolerance of its goal position and on its intended support (the floor or the cube beneath it). On the gantry, whose cubes are 16 cm, the tolerance is 7 cm in height and 8.5 cm in the plane for stacking and 7.65 cm in the plane for the line. On the Panda, whose cubes are 4 cm, it is 1.75 cm with the gripper and 1.25 cm with the pusher. Support is determined from the cube positions and not from contact forces. A cube counts as supported if its center is within a quarter of a cube edge of floor height, or if another cube lies one cube height below it, either directly underneath (within 0.6 of a cube edge horizontally) or as one of two cubes on opposite sides. A cube in an upper layer counts as placed only if the cubes it rests on in the goal structure are placed as well. Initial layouts are drawn from a pool of sampled
resting configurations, and every count $N$ is built as a scene of exactly $N$ cubes. An
episode lasts $20 N$ seconds.

\begin{figure}[h]
  \centering
  \scriptsize
  \setlength{\tabcolsep}{1pt}
  \begin{tabular}{@{}cccc@{}}
    \includegraphics[width=0.245\linewidth]{figures/snapshots/start.png} &
    \includegraphics[width=0.245\linewidth]{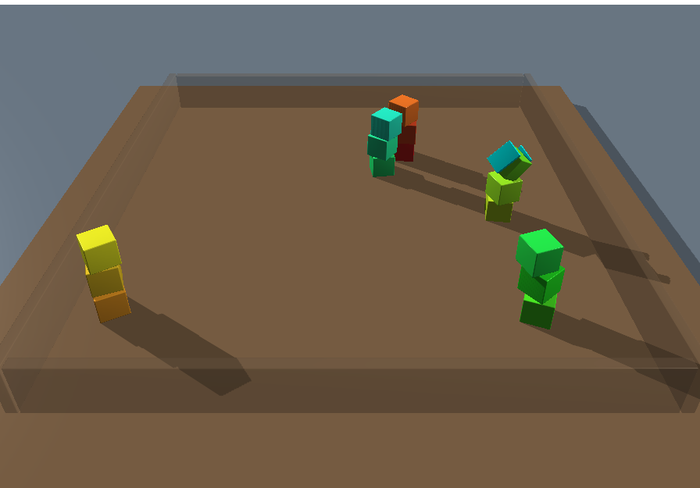} &
    \includegraphics[width=0.245\linewidth]{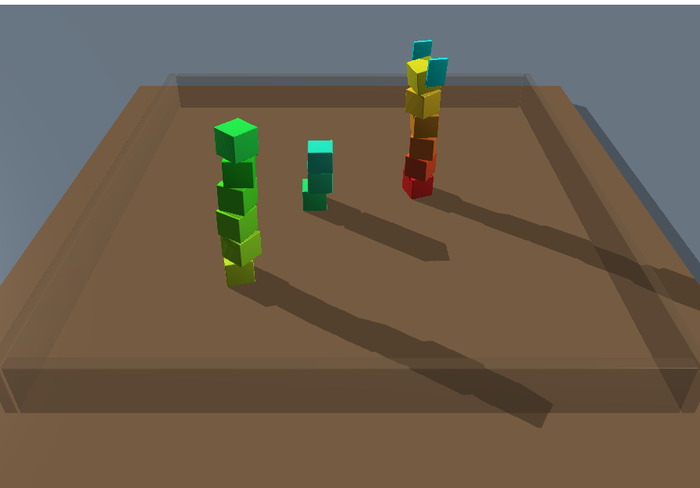} &
    \includegraphics[width=0.245\linewidth]{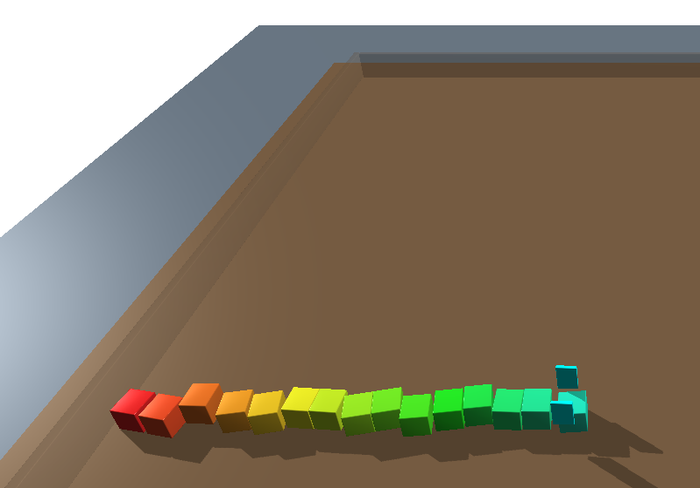} \\
    start scene & stacking, $h{=}3$ & stacking, $h{=}6$ & line \\
  \end{tabular}\\[4pt]
  \begin{tabular}{@{}ccc@{}}
    \includegraphics[width=0.245\linewidth]{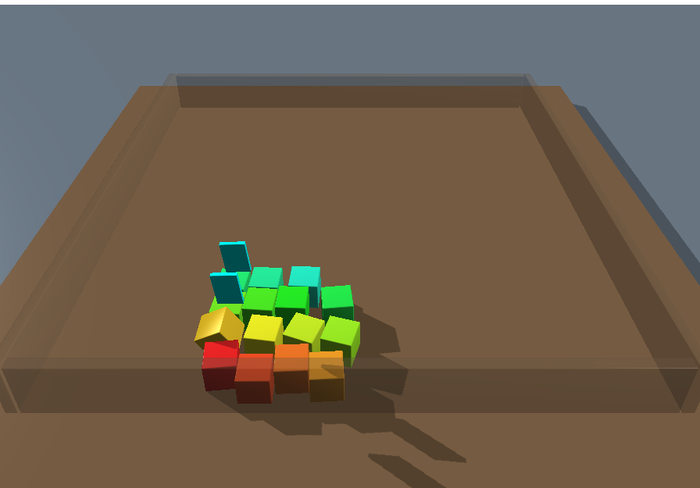} &
    \includegraphics[width=0.245\linewidth]{figures/snapshots/pyramid.png} &
    \includegraphics[width=0.245\linewidth]{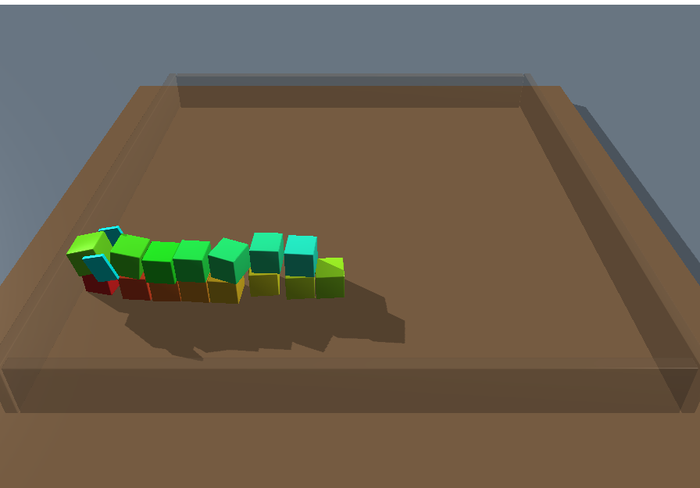} \\
    square & pyramid & wall \\
  \end{tabular}
  \caption{Renderings of the start scene and the final frames of all six gantry tasks.}
  \label{fig:app-snapshots}
\end{figure}

\begin{figure}[h]
  \centering
  \scriptsize
  \setlength{\tabcolsep}{1pt}
  \begin{tabular}{@{}ccc@{}}
    \includegraphics[width=0.30\linewidth]{figures/snapshots/panda_start.png} &
    \includegraphics[width=0.30\linewidth]{figures/snapshots/panda_stacking_h3.png} &
    \includegraphics[width=0.30\linewidth]{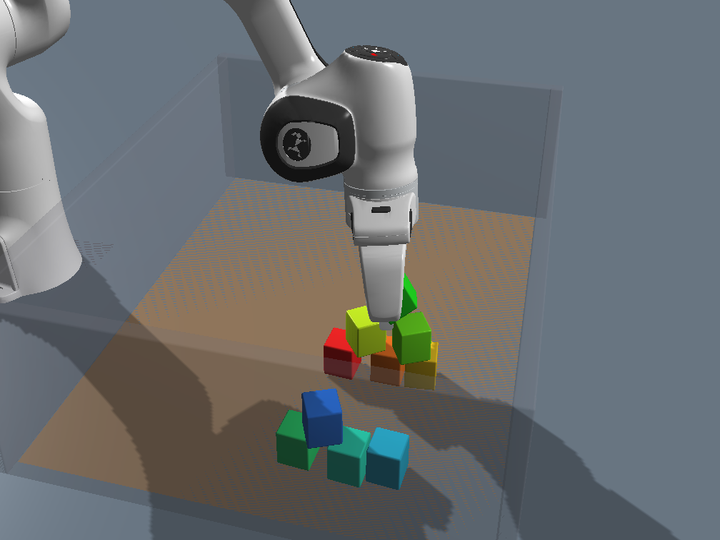} \\
    start scene & stacking, $h{=}3$ & pyramid \\
  \end{tabular}\\[4pt]
  \begin{tabular}{@{}ccc@{}}
    \includegraphics[width=0.30\linewidth]{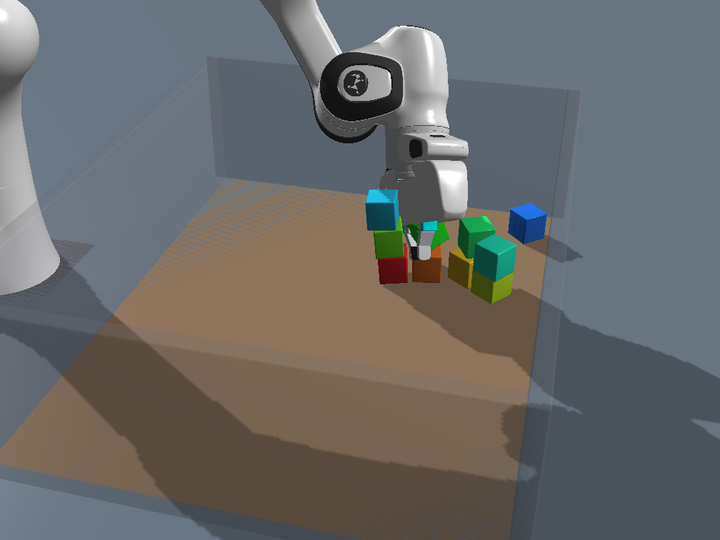} &
    \includegraphics[width=0.30\linewidth]{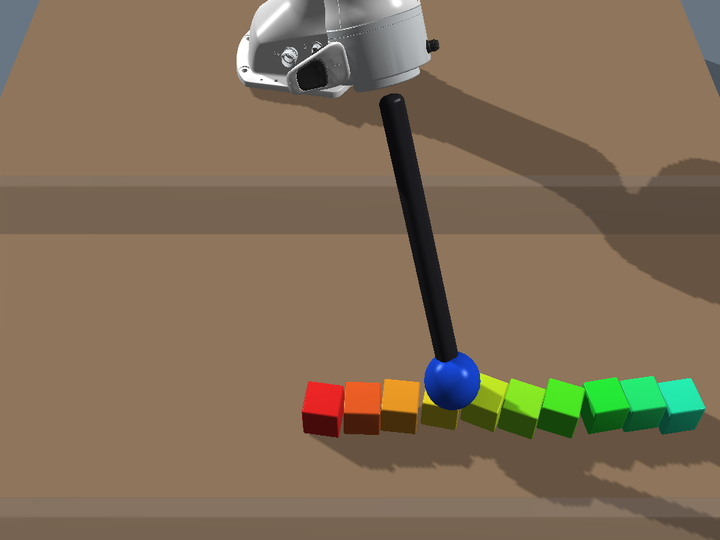} &
    \includegraphics[width=0.30\linewidth]{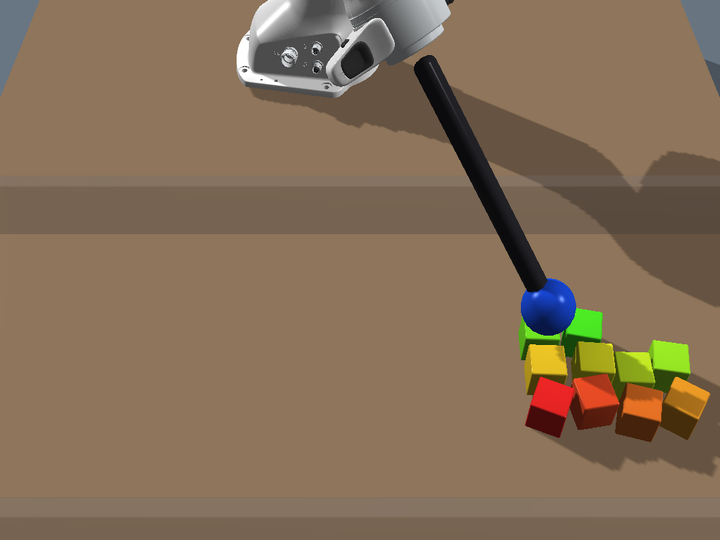} \\
    wall (9 of 10 placed) & line & square \\
  \end{tabular}
  \caption{Renderings of the start scene and the final frames of all five Panda tasks.}
  \label{fig:app-panda-snapshots}
\end{figure}

\emph{Training.} For every task we train one warping policy and one selection policy with
PPO on scenes whose object count is drawn uniformly from $1$ to $\Ntrain$, with $\Ntrain = 7$ for
cube stacking with either height, $6$ for the line and the pyramid, $9$ for the square and $8$ for the wall on the
gantry, and $7$ for cube stacking, $6$ for the pyramid, $8$ for the wall and $9$ for the line
and the square on the Panda. A gantry run uses
$5.0 \times 10^8$ environment steps (2048 environments $\times$ 96 steps $\times$ 2543
iterations) on one A100 GPU within 10 h, and every Panda run uses
$7.9 \times 10^8$ (4000 iterations) environment steps within
18 h. The warping policy uses a two
layer transformer encoder with width 128 and four heads, followed by a two layer transformer
decoder with one query per channel group that emits one categorical distribution per warp
parameter. The warp edits 55 channels of $\mu$ with
one multiplier and one offset each, i.e., 110 parameters. The 47 input channels are the
presented goal position and orientation (3 + 3), the object position and orientation (3 + 3),
the object linear and angular velocity (3 + 3), the robot joint positions and velocities
(9 + 9), the object to goal offset (3) and the servo errors (8). The 8 output channels are
the 7 motion commands and the grip command. The remaining inputs of $\mu$, such as the cube
corners, are computed from the warped channels. The channel counts fit the Panda gripper, the
largest of our robots, and on the gantry the 4 surplus channels stay at the identity. Multipliers take values on the 25
point grid of Figure~\ref{fig:warper}, and offsets take values on a 21 point grid of multiples
of a 1.5 cm step, up to 32 steps. Every parameter distribution is initialized
with 80\% of its mass on the identity value. PPO uses a learning rate of $10^{-3}$ for the
warping policy and $3 \times 10^{-4}$ for the selection policy, both annealed by a cosine schedule to a tenth, a clip ratio of 0.2,
entropy coefficients of $5 \times 10^{-4}$ and $3 \times 10^{-3}$, GAE with $\lambda_{\mathrm{GAE}} = 0.95$, and
four epochs over minibatches of 8192 samples. The selection policy scores are drawn
through a softmax during training. We train three seeds for every evaluation.

\emph{Reward.} The task reward is sparse and requires long successful manipulation trajectories before it pays a reward.
Placed cubes can be knocked over again, so keeping an arrangement must be worth as much as
extending it. Otherwise, a fast placement that knocks down other cubes would earn more return than a slower one that keeps them in place. Episode lengths grow with $N$, so the balance between these terms must not
shift with the count. $R_N$ is built from the goal completion predicate alone. Every control step pays a reward of
$w (1 - \gamma) / 2$ times the fraction of currently placed cubes, completion pays
$w \gamma$ once, and the first placement of each cube pays $b w / N$ once, with $w = 4$ and
$b = 0.65$. This per step reward makes holding a partial arrangement valuable, so a knockdown lowers the return without a dedicated penalty term. The completion bonus keeps the actual objective dominant
in the return. The first-placement bonus makes the earliest success of every cube visible to
the policy gradient, and its division by $N$ fixes its total mass across counts. The
discount is set per episode to $\gamma = 1 - 3 / D$, with $D$ the number of control steps in the episode, so the terminal bonus is discounted by $e^{-3}$ and the ratio of per step reward to
bonus is the same at every $N$.
The same constants are used for every task in Section~\ref{sec:experiments}.

\emph{Evaluation.} Initial layouts come from a held-out pool whose seed differs from
the training pool. For every count, $N \in \{1, \dots, 30\}$ on the gantry and the Panda
pusher square, up to $20$ on the pusher line and up to $15$ on the Panda gripper tasks, we
build scenes of exactly $N$ cubes and run
the first episode of each of 20 scenes with the deterministic (argmax) warping and selection policies and the
training clock of 20 s per cube. Alongside whole-task success we record, per count, the mean peak number of
simultaneously placed cubes, the retention (placed cubes at the end divided by the peak),
the number of placed to unplaced transitions, the fraction of the episode elapsed at the
last new placement and the fraction of episodes that end by timeout.

\section{Full results}
\label{sec:app-results}

Table~\ref{tab:full-results} reports the whole-task success of WAVE on the gantry per task
and count and Table~\ref{tab:panda-results} on the Panda arm, as the mean and the range over
seeds. 

\begin{table}[h]
  \caption{Whole task success of WAVE on the gantry per task and object count, the first
  episode of each of 20 scenes per count and 20 s per cube, mean with the minimum and maximum
  over three seeds in parentheses. Cube stacking builds stacks of height $h$. The line is evaluated in a workspace widened to
  fit thirty cubes, and the wall is two courses high and grows longer with the count.}
  \label{tab:full-results}
  \begin{center}
  \fontsize{7.5}{9}\selectfont
  \begin{tabular}{lcccccc}
    \toprule
    $N$ & stacking, $h=3$ & stacking, $h=6$ & line & square & pyramid & wall \\
    \midrule
    1 &  1.00~(1.00--1.00) & 1.00~(1.00--1.00) &  1.00~(1.00--1.00) &  1.00~(1.00--1.00) &  1.00~(1.00--1.00) & 1.00~(1.00--1.00) \\
    3 &  1.00~(1.00--1.00) & 1.00~(1.00--1.00) &  1.00~(1.00--1.00) &  1.00~(1.00--1.00) &  1.00~(1.00--1.00) & 1.00~(1.00--1.00) \\
    5 &  0.95~(0.85--1.00) & 1.00~(1.00--1.00) &  1.00~(1.00--1.00) &  1.00~(1.00--1.00) &  1.00~(1.00--1.00) & 0.98~(0.95--1.00) \\
    7 &  1.00~(1.00--1.00) & 1.00~(1.00--1.00) &  1.00~(1.00--1.00) &  0.98~(0.95--1.00) &  0.92~(0.85--0.95) & 0.98~(0.95--1.00) \\
    10 &  0.98~(0.95--1.00) & 1.00~(1.00--1.00) &  1.00~(1.00--1.00) &  0.97~(0.95--1.00) &  0.90~(0.80--0.95) & 0.93~(0.85--1.00) \\
    12 &  0.98~(0.95--1.00) & 0.85~(0.80--0.95) &  0.97~(0.95--1.00) &  0.98~(0.95--1.00) &  0.80~(0.80--0.80) & 0.95~(0.90--1.00) \\
    15 &  0.92~(0.85--1.00) & 0.72~(0.55--0.85) &  1.00~(1.00--1.00) &  0.97~(0.90--1.00) &  0.60~(0.45--0.70) & 0.85~(0.70--0.95) \\
    20 &  0.95~(0.90--1.00) & 0.13~(0.10--0.20) &  0.93~(0.90--1.00) &  0.85~(0.65--0.95) &  0.35~(0.20--0.45) & 0.67~(0.45--0.80) \\
    25 &  0.63~(0.45--0.80) & 0.00~(0.00--0.00) &  0.97~(0.95--1.00) &  0.77~(0.65--0.85) &  0.20~(0.15--0.25) & 0.63~(0.40--0.75) \\
    28 &  0.53~(0.35--0.65) & 0.00~(0.00--0.00) &  0.92~(0.85--0.95) &  0.80~(0.60--1.00) &  0.10~(0.05--0.15) & 0.48~(0.10--0.75) \\
    30 &  0.48~(0.30--0.60) & 0.00~(0.00--0.00) &  0.90~(0.80--1.00) &  0.55~(0.45--0.60) &  0.08~(0.05--0.15) & 0.38~(0.05--0.85) \\
    \bottomrule
  \end{tabular}
  \end{center}
\end{table}

\begin{table}[h]
  \caption{Whole task success of WAVE on the Panda arm per task and object count, the same
  protocol as Table~\ref{tab:full-results}. Cube stacking ($h=3$), the pyramid and the wall use the
  parallel-jaw gripper and the line and the square the sphere-tip pusher. A dash marks a
  count the robot was not evaluated on.}
  \label{tab:panda-results}
  \begin{center}
  \fontsize{7.5}{9}\selectfont
  \begin{tabular}{lccccc}
    \toprule
    & \multicolumn{3}{c}{gripper} & \multicolumn{2}{c}{pusher} \\
    \cmidrule(lr){2-4} \cmidrule(lr){5-6}
    $N$ & stacking, $h=3$ & pyramid & wall & line & square \\
    \midrule
    1  & 1.00~(1.00--1.00) & 1.00~(1.00--1.00) & 0.98~(0.95--1.00) & 1.00~(1.00--1.00) & 1.00~(1.00--1.00) \\
    3  & 0.92~(0.85--1.00) & 0.90~(0.85--0.95) & 0.70~(0.65--0.75) & 1.00~(1.00--1.00) & 1.00~(1.00--1.00) \\
    5  & 0.87~(0.75--0.95) & 0.47~(0.15--0.65) & 0.23~(0.10--0.40) & 1.00~(1.00--1.00) & 1.00~(1.00--1.00) \\
    7  & 0.63~(0.60--0.70) & 0.28~(0.10--0.55) & 0.03~(0.00--0.10) & 0.98~(0.95--1.00) & 0.92~(0.75--1.00) \\
    9  & 0.38~(0.20--0.60) & 0.17~(0.10--0.20) & 0.00~(0.00--0.00) & 0.95~(0.90--1.00) & 0.60~(0.50--0.70) \\
    12 & 0.03~(0.00--0.10) & 0.00~(0.00--0.00) & 0.00~(0.00--0.00) & 0.80~(0.75--0.85) & 0.35~(0.20--0.55) \\
    15 & 0.03~(0.00--0.10) & 0.00~(0.00--0.00) & 0.00~(0.00--0.00) & 0.78~(0.75--0.80) & 0.10~(0.00--0.15) \\
    20 & -- & -- & -- & 0.62~(0.50--0.70) & 0.00~(0.00--0.00) \\
    25 & -- & -- & -- & -- & 0.00~(0.00--0.00) \\
    30 & -- & -- & -- & -- & 0.00 \\
    \bottomrule
  \end{tabular}
  \end{center}
\end{table}

\begin{figure}[h]
  \centering
  \begin{minipage}[t]{0.48\linewidth}
    \vspace{0pt}
    \centering
    line
    \includegraphics[width=\linewidth]{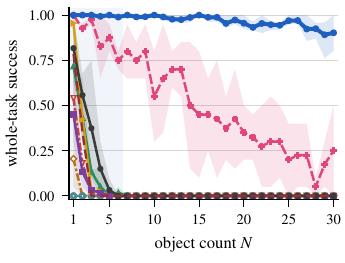}\\[1pt]
  \end{minipage}\hfill
  \begin{minipage}[t]{0.48\linewidth}
    \vspace{0pt}
    \centering
    cube stacking, $h=3$
    \includegraphics[width=\linewidth]{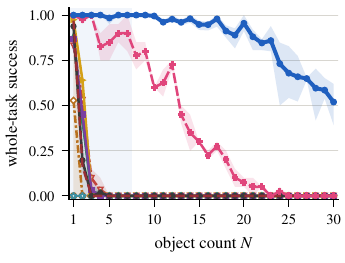}\\[1pt]
  \end{minipage}
  \\[2pt]
  \includegraphics[width=0.85\linewidth]{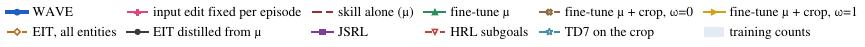}
  \caption{Whole task success for WAVE and all ten baselines of Section~\ref{sec:exp-baselines}, left on the line and right on cube stacking. The curves are the mean over three seeds and the bands the seed range.}
  \label{fig:snapshots}
\end{figure}

\begin{table}[h]
  \caption{Whole task success per object count for WAVE and the baselines of
  Section~\ref{sec:exp-baselines} (Figure~\ref{fig:snapshots}), the mean over three seeds with the minimum and maximum over the seeds in parentheses below it.
  Every method trains on one to six cubes on the line and one to seven on cube stacking.
  Bold marks the highest mean per task and count.}
  \label{tab:baselines}
  \begin{center}
  \fontsize{7}{8.5}\selectfont
  \setlength{\tabcolsep}{3pt}
  \begin{tabular}{lcccccccc}
    \toprule
    $N$ & 1 & 2 & 3 & 4 & 5 & 6 & 10 & 30 \\
    \midrule
    \multicolumn{9}{l}{\emph{line}} \\
    WAVE & \textbf{1.00} & \textbf{1.00} & \textbf{1.00} & \textbf{0.98} & \textbf{1.00} & \textbf{1.00} & \textbf{1.00} & \textbf{0.90} \\
    & (1.00--1.00) & (1.00--1.00) & (1.00--1.00) & (0.95--1.00) & (1.00--1.00) & (1.00--1.00) & (1.00--1.00) & (0.80--1.00) \\[1pt]
    input edit fixed per episode & \textbf{1.00} & 0.93 & 0.98 & 0.83 & 0.88 & 0.75 & 0.55 & 0.25 \\
    & (1.00--1.00) & (0.90--0.95) & (0.95--1.00) & (0.65--1.00) & (0.80--0.95) & (0.70--0.80) & (0.30--0.80) & (0.00--0.50) \\[1pt]
    skill alone ($\mu$) & 0.79 & 0.35 & 0.00 & 0.00 & 0.00 & 0.00 & 0.00 & 0.00 \\
    & (0.75--0.83) & (0.29--0.43) & (0.00--0.00) & (0.00--0.00) & (0.00--0.00) & (0.00--0.00) & (0.00--0.00) & (0.00--0.00) \\[1pt]
        fine-tuning $\mu$ & 0.72 & 0.43 & 0.13 & 0.05 & 0.00 & 0.02 & 0.00 & 0.00 \\
    & (0.65--0.75) & (0.30--0.65) & (0.10--0.15) & (0.00--0.15) & (0.00--0.00) & (0.00--0.05) & (0.00--0.00) & (0.00--0.00) \\[1pt]
    \quad on the crop, $\omega=0$ & 0.00 & 0.00 & 0.00 & 0.00 & 0.00 & 0.00 & 0.00 & 0.00 \\
    & (0.00--0.00) & (0.00--0.00) & (0.00--0.00) & (0.00--0.00) & (0.00--0.00) & (0.00--0.00) & (0.00--0.00) & (0.00--0.00) \\[1pt]
    \quad on the crop, $\omega=1$ & 0.96 & 0.43 & 0.03 & 0.03 & 0.00 & 0.00 & 0.00 & 0.00 \\
    & (0.93--0.98) & (0.40--0.45) & (0.00--0.05) & (0.00--0.05) & (0.00--0.00) & (0.00--0.00) & (0.00--0.00) & (0.00--0.00) \\[1pt]
    EIT~\citep{haramati2024} & 0.20 & 0.00 & 0.00 & 0.00 & 0.00 & 0.00 & 0.00 & 0.00 \\
    & (0.05--0.36) & (0.00--0.00) & (0.00--0.00) & (0.00--0.00) & (0.00--0.00) & (0.00--0.00) & (0.00--0.00) & (0.00--0.00) \\[1pt]
    EIT distilled from $\mu$ & 0.82 & 0.56 & 0.37 & 0.15 & 0.03 & 0.00 & 0.00 & 0.00 \\
    & (0.62--0.99) & (0.43--0.68) & (0.10--0.75) & (0.05--0.30) & (0.00--0.10) & (0.00--0.00) & (0.00--0.00) & (0.00--0.00) \\[1pt]
        JSRL~\citep{uchendu2022} & 0.45 & 0.13 & 0.03 & 0.02 & 0.00 & 0.00 & 0.00 & 0.00 \\
    & (0.30--0.65) & (0.05--0.20) & (0.00--0.10) & (0.00--0.05) & (0.00--0.00) & (0.00--0.00) & (0.00--0.00) & (0.00--0.00) \\[1pt]
    HRL subgoals & 0.54 & 0.18 & 0.00 & 0.00 & 0.00 & 0.00 & 0.00 & 0.00 \\
    & (0.35--0.73) & (0.15--0.20) & (0.00--0.00) & (0.00--0.00) & (0.00--0.00) & (0.00--0.00) & (0.00--0.00) & (0.00--0.00) \\[1pt]
    TD7 on the crop & 0.00 & 0.00 & 0.00 & 0.00 & 0.00 & 0.00 & 0.00 & 0.00 \\
    & (0.00--0.00) & (0.00--0.00) & (0.00--0.00) & (0.00--0.00) & (0.00--0.00) & (0.00--0.00) & (0.00--0.00) & (0.00--0.00) \\[1pt]
    \midrule
    \multicolumn{9}{l}{\emph{cube stacking, $h=3$}} \\
    WAVE & \textbf{1.00} & \textbf{1.00} & \textbf{1.00} & \textbf{1.00} & \textbf{0.95} & \textbf{1.00} & \textbf{0.98} & \textbf{0.48} \\
    & (1.00--1.00) & (1.00--1.00) & (1.00--1.00) & (1.00--1.00) & (0.85--1.00) & (1.00--1.00) & (0.95--1.00) & (0.30--0.60) \\[1pt]
    input edit fixed per episode & \textbf{1.00} & 0.98 & \textbf{1.00} & 0.83 & 0.85 & 0.90 & 0.60 & 0.00 \\
    & (1.00--1.00) & (0.95--1.00) & (1.00--1.00) & (0.75--0.90) & (0.75--0.95) & (0.85--0.95) & (0.60--0.60) & (0.00--0.00) \\[1pt]
    skill alone ($\mu$) & 0.97 & 0.17 & 0.00 & 0.00 & 0.00 & 0.00 & 0.00 & 0.00 \\
    & (0.95--0.99) & (0.13--0.20) & (0.00--0.00) & (0.00--0.00) & (0.00--0.00) & (0.00--0.00) & (0.00--0.00) & (0.00--0.00) \\[1pt]
        fine-tuning $\mu$ & 0.88 & 0.42 & 0.02 & 0.02 & 0.00 & 0.00 & 0.00 & 0.00 \\
    & (0.85--0.95) & (0.35--0.50) & (0.00--0.05) & (0.00--0.05) & (0.00--0.00) & (0.00--0.00) & (0.00--0.00) & (0.00--0.00) \\[1pt]
    \quad on the crop, $\omega=0$ & 0.00 & 0.00 & 0.00 & 0.00 & 0.00 & 0.00 & 0.00 & 0.00 \\
    & (0.00--0.00) & (0.00--0.00) & (0.00--0.00) & (0.00--0.00) & (0.00--0.00) & (0.00--0.00) & (0.00--0.00) & (0.00--0.00) \\[1pt]
    \quad on the crop, $\omega=1$ & 0.98 & 0.64 & 0.03 & 0.00 & 0.00 & 0.00 & 0.00 & 0.00 \\
    & (0.95--1.00) & (0.61--0.67) & (0.00--0.05) & (0.00--0.00) & (0.00--0.00) & (0.00--0.00) & (0.00--0.00) & (0.00--0.00) \\[1pt]
    EIT~\citep{haramati2024} & 0.53 & 0.00 & 0.00 & 0.00 & 0.00 & 0.00 & 0.00 & 0.00 \\
    & (0.00--0.90) & (0.00--0.00) & (0.00--0.00) & (0.00--0.00) & (0.00--0.00) & (0.00--0.00) & (0.00--0.00) & (0.00--0.00) \\[1pt]
    EIT distilled from $\mu$ & 0.94 & 0.20 & 0.00 & 0.02 & 0.00 & 0.00 & 0.00 & 0.00 \\
    & (0.85--0.99) & (0.00--0.32) & (0.00--0.00) & (0.00--0.05) & (0.00--0.00) & (0.00--0.00) & (0.00--0.00) & (0.00--0.00) \\[1pt]
        JSRL~\citep{uchendu2022} & 0.87 & 0.45 & 0.03 & 0.00 & 0.00 & 0.00 & 0.00 & 0.00 \\
    & (0.85--0.90) & (0.35--0.60) & (0.00--0.10) & (0.00--0.00) & (0.00--0.00) & (0.00--0.00) & (0.00--0.00) & (0.00--0.00) \\[1pt]
    HRL subgoals & 0.83 & 0.53 & 0.10 & 0.03 & 0.00 & 0.00 & 0.00 & 0.00 \\
    & (0.77--0.91) & (0.43--0.59) & (0.05--0.15) & (0.00--0.10) & (0.00--0.00) & (0.00--0.00) & (0.00--0.00) & (0.00--0.00) \\[1pt]
    TD7 on the crop & 0.00 & 0.00 & 0.00 & 0.00 & 0.00 & 0.00 & 0.00 & 0.00 \\
    & (0.00--0.00) & (0.00--0.00) & (0.00--0.00) & (0.00--0.00) & (0.00--0.00) & (0.00--0.00) & (0.00--0.00) & (0.00--0.00) \\[1pt]
    \bottomrule
  \end{tabular}
  \end{center}
\end{table}

\begin{table}[h]
  \caption{How the episodes of WAVE on the gantry end, per task and object count, with
  whole-task success, at the time limit, or because a cube left the workspace. The policies and
  the protocol are those of Table~\ref{tab:full-results} but evaluated in a separate run with unpinned stochasticities, so the success rows agree with that
  table within its sampling error. For each cause of episode end we give the 
  mean, the minimum and maximum over three seeds. No episode ended with a
  simulation failure.}
  \label{tab:end-causes}
  \begin{center}
  \fontsize{7.5}{9}\selectfont
  \setlength{\tabcolsep}{3.5pt}
  \begin{tabular}{@{}rl@{\hspace{7pt}}cccccc@{}}
    \toprule
    $N$ & ends with & stacking, $h=3$ & stacking, $h=6$ & line & square & pyramid & wall \\
    \midrule
    5 & success & 0.98~(0.97--1.00) & 0.99~(0.98--1.00) & 1.00~(1.00--1.00) & 0.99~(0.97--1.00) & 0.99~(0.97--1.00) & 0.99~(0.98--1.00) \\
     & time limit & 0.01~(0.00--0.02) & 0.00~(0.00--0.00) & 0.00~(0.00--0.00) & 0.00~(0.00--0.00) & 0.01~(0.00--0.03) & 0.01~(0.00--0.02) \\
     & cube lost & 0.01~(0.00--0.02) & 0.01~(0.00--0.02) & 0.00~(0.00--0.00) & 0.01~(0.00--0.03) & 0.00~(0.00--0.00) & 0.00~(0.00--0.00) \\
    \addlinespace
    10 & success & 0.94~(0.86--1.00) & 1.00~(1.00--1.00) & 0.99~(0.98--1.00) & 0.98~(0.97--1.00) & 0.91~(0.88--0.97) & 0.97~(0.95--1.00) \\
     & time limit & 0.05~(0.00--0.12) & 0.00~(0.00--0.00) & 0.01~(0.00--0.02) & 0.01~(0.00--0.02) & 0.09~(0.03--0.12) & 0.02~(0.00--0.05) \\
     & cube lost & 0.01~(0.00--0.02) & 0.00~(0.00--0.00) & 0.01~(0.00--0.02) & 0.02~(0.00--0.03) & 0.00~(0.00--0.00) & 0.01~(0.00--0.02) \\
    \addlinespace
    15 & success & 0.90~(0.84--0.92) & 0.86~(0.84--0.88) & 0.97~(0.94--1.00) & 0.95~(0.92--0.98) & 0.61~(0.56--0.66) & 0.92~(0.83--0.97) \\
     & time limit & 0.09~(0.06--0.14) & 0.10~(0.08--0.12) & 0.02~(0.00--0.03) & 0.05~(0.02--0.08) & 0.36~(0.33--0.41) & 0.07~(0.00--0.17) \\
     & cube lost & 0.01~(0.00--0.02) & 0.04~(0.03--0.05) & 0.01~(0.00--0.03) & 0.00~(0.00--0.00) & 0.03~(0.02--0.03) & 0.02~(0.00--0.03) \\
    \addlinespace
    20 & success & 0.81~(0.75--0.86) & 0.13~(0.08--0.20) & 0.96~(0.92--0.98) & 0.88~(0.80--0.92) & 0.43~(0.30--0.50) & 0.77~(0.56--0.89) \\
     & time limit & 0.14~(0.09--0.22) & 0.74~(0.69--0.77) & 0.03~(0.02--0.05) & 0.09~(0.05--0.16) & 0.53~(0.45--0.67) & 0.20~(0.06--0.41) \\
     & cube lost & 0.06~(0.03--0.09) & 0.13~(0.11--0.16) & 0.01~(0.00--0.03) & 0.03~(0.02--0.05) & 0.04~(0.03--0.05) & 0.03~(0.02--0.05) \\
    \addlinespace
    25 & success & 0.66~(0.59--0.77) & 0.00~(0.00--0.00) & 0.93~(0.92--0.94) & 0.77~(0.70--0.81) & 0.18~(0.11--0.22) & 0.60~(0.12--0.97) \\
     & time limit & 0.26~(0.19--0.33) & 0.85~(0.83--0.88) & 0.04~(0.02--0.06) & 0.20~(0.12--0.30) & 0.80~(0.77--0.88) & 0.40~(0.03--0.88) \\
     & cube lost & 0.09~(0.05--0.14) & 0.15~(0.12--0.17) & 0.03~(0.02--0.05) & 0.04~(0.00--0.06) & 0.02~(0.02--0.03) & 0.01~(0.00--0.02) \\
    \addlinespace
    30 & success & 0.45~(0.39--0.48) & 0.00~(0.00--0.00) & 0.92~(0.88--0.95) & 0.66~(0.66--0.67) & 0.06~(0.00--0.16) & 0.40~(0.08--0.69) \\
     & time limit & 0.49~(0.44--0.58) & 0.76~(0.72--0.81) & 0.05~(0.02--0.09) & 0.28~(0.25--0.30) & 0.90~(0.81--0.95) & 0.56~(0.25--0.89) \\
     & cube lost & 0.06~(0.03--0.08) & 0.24~(0.19--0.28) & 0.03~(0.03--0.03) & 0.06~(0.05--0.08) & 0.04~(0.03--0.05) & 0.04~(0.02--0.06) \\
    \bottomrule
  \end{tabular}
  \end{center}
\end{table}

\section{Real-world Panda experiments}
\label{sec:app-realworld}

We deployed a cube stacking policy trained in simulation on a physical Panda arm. The policies are state-based, so we estimate the poses of the $4 \times 4 \times 4$\,cm cubes from AprilTag fiducials tracked by four calibrated RGB cameras.

\begin{figure}[t]
\centering
\includegraphics[width=\textwidth]{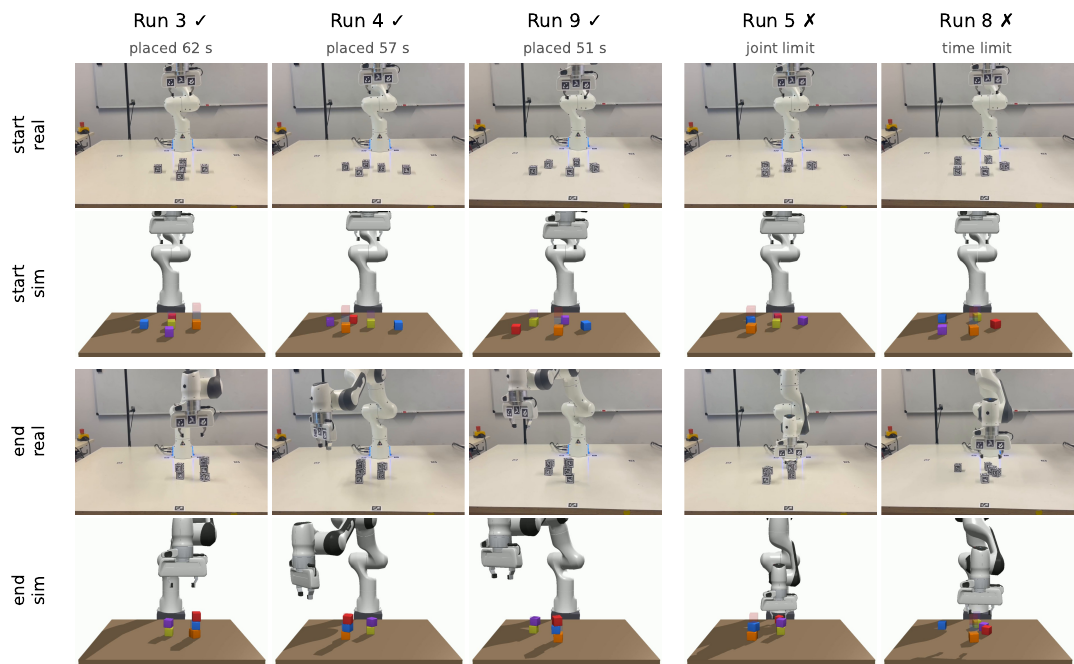}
\caption{Selected real rollouts of building a three-cube and a two-cube tower (columns) with the same
rollouts replayed in simulation from the logged joint states, gripper widths and tracked cube poses.
The rows show the start and end on the real robot and in the replay. Runs 3, 4 and 9 succeeded. Run 5 stopped
at a joint limit and run 8 at the time limit.}
\label{fig:real_series}
\end{figure}

We evaluate the policy on ten consecutive rollouts of building a three-cube and a two-cube tower from
five cubes (Table~\ref{tab:real_series}, Figure~\ref{fig:real_series}). The two base cubes start in
their goal slots, so three cubes have to be picked and placed. The initial layout differs between
runs. In 8 of the 10 runs all five cubes reached their goals (95\% Wilson interval 49--94\%), after
$410 \pm 248$ policy steps ($92 \pm 54$\,s) on average. In 5 of the 10 runs the policy also released
the last cube and both towers stood at the end. The gap between the two rates comes from the final
release, which the policy never learned. In training and in the simulation evaluation, an episode ends as a success as soon as every cube is within tolerance and on its support, even if the last cube is still in the gripper. Appendix~\ref{sec:app-generalization} reports in how many of these episodes every cube is still placed 2\,s after a scripted release. On the robot the rollout
continues, and without a remaining objective the policy keeps the cube near its goal (runs 6, 7 and
10) or lets go only after a long delay (29\,s in run 2, about two minutes in run 3). Requiring the
release for success during training would remove this failure mode, but we leave this for future work. The two failed runs were stopped at a
joint limit with four of five cubes placed (run 5) and at the time limit (run 8).
Grasps often needed several attempts. Of the 78 close
commands, 68\% held a cube, and the policy recovered from failed grasps by trying again. Runs 4 and 9
needed the minimum of three grasps. Runs 6, 7 and 10 needed 9--11. The policy releases a cube a
median of 13\,mm above its goal slot, a behavior learned in simulation, where the jaws open slowly.
Together with gripper stiction on the real system, this was the main reason cubes toppled.

\begin{figure}[t]
  \centering
  \includegraphics[width=\textwidth]{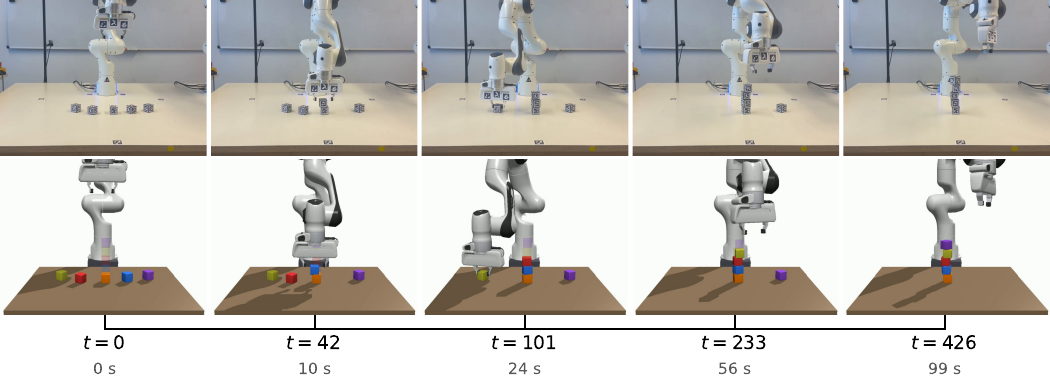}
  \caption{Five-high tower built on the real robot.}
  \label{fig:tower5_real}
\end{figure}

We also deployed a policy fine-tuned for five-cube towers. It built a five-high tower on the real robot (Figure~\ref{fig:tower5_real}), but we did not evaluate this task systematically.

State estimation added a further source of error. We applied no temporal filtering, so tracking jitter, occluded fiducials and camera calibration errors entered the observation of the policy directly. The real gripper also has velocity limits and more stiction than the simulated one. A motion-capture system with temporal filtering and domain randomization of gripper friction and stiction would address both sources of error.

\begin{table}[t]
  \centering
  \small
  \setlength{\tabcolsep}{4pt}
  
  \caption{Ten consecutive real-robot rollouts of the task of building a three-cube and a two-cube tower from five cubes (two base cubes start in their slots, three have to be moved). \emph{Placed} means all five cubes are within tolerance of their goals. \emph{Released} additionally requires that the last cube was let go and the structure stands at the end of the run. Steps are policy actions at 5\,Hz, with the time in seconds in parentheses and means over the successful runs. \emph{Grasps} counts grasps that held a cube, \emph{closes} counts gripper close commands and \emph{cubes} counts cubes within tolerance at the end.}
  \begin{tabular}{rccrrcccl}
    \toprule
    & \multicolumn{2}{c}{success} & \multicolumn{2}{c}{steps to (time [s])} & & & & \\
    \cmidrule(lr){2-3} \cmidrule(lr){4-5}
    run & placed & released & placed & released & grasps & closes & cubes & stop \\
    \midrule
    1 & \checkmark & \checkmark & 373 (82) & 472 (102) & 5 & 5 & 5/5 &  \\
    2 & \checkmark & \checkmark & 235 (53) & 402 (88) & 4 & 5 & 5/5 &  \\
    3 & \checkmark & \checkmark & 276 (62) & 873 (184) & 5 & 5 & 5/5 &  \\
    4 & \checkmark & \checkmark & 254 (57) & 259 (58) & 3 & 4 & 5/5 &  \\
    5 & -- & -- & -- & -- & 2 & 8 & 4/5 & joint limit \\
    6 & \checkmark & -- & 300 (66) & -- & 9 & 11 & 5/5 &  \\
    7 & \checkmark & -- & 844 (194) & -- & 9 & 16 & 5/5 &  \\
    8 & -- & -- & -- & -- & 2 & 6 & 2/5 & time limit \\
    9 & \checkmark & \checkmark & 234 (51) & 252 (55) & 3 & 3 & 5/5 &  \\
    10 & \checkmark & -- & 764 (175) & -- & 11 & 15 & 4/5 &  \\
    \midrule
    rate & 8/10 & 5/10 & & & & & & \\
    95\% CI & 49--94\% & 24--76\% & & & & & & \\
    mean & & & 410 $\pm$ 248 & 452 $\pm$ 254 & 5.3 & 7.8 & 4.5/5 & \\
    \bottomrule
  \end{tabular}
  \label{tab:real_series}
\end{table}

\section{Baselines}
\label{sec:app-baselines}

Every baseline of Section~\ref{sec:exp-baselines} runs on the same gantry and task as WAVE,
with the same reward unless noted, the same mixed training counts (one to six cubes on the
line, one to seven on cube stacking) and $5 \times 10^8$ environment steps over 2048 parallel scenes,
and every baseline that uses a skill uses the same checkpoint of $\mu$. They are evaluated like
WAVE from one to thirty cubes over three seeds. 

We supply a scripted selection policy that chooses only objects with goals that are currently fulfillable. For example, in cube stacking only cubes in the bottom layer or on an already placed cube can be selected. The selection then chooses the fulfillable object nearest to the gripper. WAVE does not use a scripted selection policy, so the baselines train on an easier task. All baselines use this selection policy, except for the EIT baselines, HRL subgoals and the fixed input edit. The last two use the learned selection policy of WAVE.

Table~\ref{tab:baselines} compares WAVE with the baselines of
Section~\ref{sec:exp-baselines} on the line and on cube stacking ($h=3$), with the mean and the range over all seeds. Table~\ref{tab:exp-baselines} in Section~\ref{sec:exp-baselines} shows seven of these baselines. Plain
fine-tuning of $\mu$, fine-tuning on the crop without the anchor ($\omega = 0$) and JSRL are reported only here, in
Table~\ref{tab:baselines} and Figure~\ref{fig:snapshots}.

\emph{Skill alone} ($\mu$) runs the frozen skill with no training on the selected object. \emph{Fine-tuning} $\mu$ keeps the same selection but continues to train $\mu$'s own actor and critic on the multi-object reward. During the first $10\%$ of training only the critic is updated and the actor remains equal to $\mu$, so that the critic can adapt to the multi-object reward before the actor changes.
\emph{EIT} is the Entity Interaction Transformer of \citet{haramati2024}
with every object and goal as a token, trained with TD3 and hindsight relabeling, with the same hyperparameters as reported in their paper. 
\emph{EIT distilled from} $\mu$ first imitates the actions of $\mu$, whose
rollouts also fill its replay buffer. It is then trained like \emph{EIT}, with an added loss that keeps the actor close to $\mu$. In \emph{JSRL}~\citep{uchendu2022} $\mu$ is used as a guide policy that
acts the first $90\%$ of every training episode and a TD7 exploration policy initialized
from $\mu$'s weights acts the rest of the episode. Control is handed over to the exploration policy in stages, each time the success of the combined policy stops improving, until the exploration policy alone acts on the robot at the end of training. \emph{HRL subgoals} is a hierarchy with the
frozen $\mu$ as its low level, which keeps WAVE's selection policy and its
crop $\chi$, and in place of the warp a TD7 manager that reads the crop and
proposes goals for $\mu$ to pursue. \emph{TD7 on the crop} trains a TD7 policy from random
initialization on the same crop $\chi$ as WAVE, with the scripted selection and without $\mu$. \emph{Input edit fixed per episode} is WAVE with one modification: the input side of the warp is chosen at the first step of an episode and kept until the episode ends, as the state projection of \citet{zhang2023b} is, while the output side is still chosen at every step. Offsets, crop, selection policy and training budget are those of WAVE.

\emph{Fine-tuning $\mu$ on the crop} gives the fine-tuned $\mu$ the view of the other objects that plain fine-tuning lacks.
The crop $\chi$ is the same as for the warping policy (Section~\ref{sec:warper}). This
crop is passed to the first layer of $\mu$'s actor as additional input, through new weights initialized at zero, so that the actor outputs exactly $\mu(\sigma)$ at the start of training. The critic is extended in the same way and reads $(\sigma, \chi, a)$. For the first $10\%$ of training only the critic is trained and the actor stays
fixed, so that the critic has learned the multi-object reward before it updates the actor. The actor then minimizes the TD7
loss with an anchor to the frozen skill, as in TD3+BC~\citep{fujimoto2021},
\begin{equation*}
  \mathcal{L}(\theta) = -\frac{Q(\sigma, \chi, \pi_\theta(\sigma, \chi))}{\overline{|Q|}}
  + \omega \, \bigl\lVert \pi_\theta(\sigma, \chi) - \mu(\sigma) \bigr\rVert^2 ,
\end{equation*}
where $\overline{|Q|}$ is the mean absolute value of $Q$ over the batch, which makes $\omega$ independent of the scale of the
reward. With $\omega = 0$ the actor is trained on the reward alone. With $\omega = 1$, staying close to the action of $\mu$ is
weighted about as much as improving the normalized value. This baseline keeps the scripted selection, and like $\mu$ inside WAVE it
chooses an action at every control step.

\section{Overview of published methods and results}
\label{sec:app-published}

\begin{table}[h!]
  \caption{Training conditions and best reported results of the published methods cited in
  Section~\ref{sec:related} that learn the manipulation itself, in a physics simulator or on a
  real robot, with a goal pose for every object. All values are taken from their reports. The
  gray series of Figure~\ref{fig:results} are the rows that report whole-task success per
  count.}
  \label{tab:published-learned}
  \begin{center}
  \fontsize{7.5}{9}\selectfont
  \renewcommand{\arraystretch}{1.15}
  \begin{tabular}{@{}>{\raggedright\arraybackslash}p{1.9cm}>{\raggedright\arraybackslash}p{3.1cm}>{\raggedright\arraybackslash}p{2.0cm}>{\raggedright\arraybackslash}p{1.9cm}>{\raggedright\arraybackslash}p{3.4cm}@{}}
    \toprule
    \textit{Method} & \textit{Trains on} & \textit{Task} &
    \textit{Counts (train $\to$ test)} & \textit{Best reported} \\
    \midrule
    \citet{li2020} & MuJoCo Fetch, privileged state, attention graph network &
      stacking every block & curriculum 1--6 $\to$ 9 & 0.93 at 4, 0.84 at 5, 0.75 at 6, separate policies per count \\
    ECRL~\citep{haramati2024} & IsaacGym, two RGB views to latent particles, end effector position deltas
      on a pusher & planar pushing, 2-D goals & 3 $\to$ 1--6 &
      0.84 at 3, 0.40 at 6, and a 12 object sort is shown qualitatively only \\
    HECRL~\citep{haramati2026} & MuJoCo adapted from OGBench, state and image variants &
      rearrangement, stacking & 3 $\to$ 1--6 (stacking 2--4) &
      0.83 at 3 and 0.26 at 6 on rearrangement from state, 0.64 and 0.10 from images \\
    EC-Diffuser~\citep{ecdiffuser2025} & IsaacGym, two RGB views to latent particles, behavior
      cloning & pushing cubes and T blocks & 3 $\to$ 6 & 0.24 at 6, 0.83 at 3 by pushing cubes \\
    SMORL~\citep{zadaianchuk2021} & MuJoCo, RGB to object slots & pushing and rearranging
      pucks & per count, 1--2 from pixels and 2--4 from state & --- (transfer only downward, 2
      to 1) \\
    PAIR~\citep{li2022} & PyBullet, privileged state, discretized actions & six-cube stacking &
      6 $\to$ 6 & 0.955 at 6, where success counts one target cube among six \\
    \citet{lanier2019} & MuJoCo Fetch, privileged state, binary or incremental reward &
      stacking & per count, 2--4 & 0.98 at 3 and 0.79 at 4 with the incremental reward, 0.95 and
      0.00
      at 4 with the binary one, all whole-task success \\
    LRN~\citep{mambelli2022} & CausalWorld TriFinger, privileged state & pushing and lifting
      one cube among distractors & 3 cubes $\to$ 1--10 cubes & --- (distractors, not goal
      objects) \\
    OGBench~\citep{ogbench2024} & offline datasets, state, UR5e, one policy per count &
      five goals per count, pick and place, cycles and one stack from two cubes on &
      per count, 1--4 & 0.68, 0.40, 0.04 and 0.00 at 1 to 4 over the five goals; on the
      stack goal 0.23 at 2 and 0.00 at 3 and 4, best of six baselines \\
    SHARSA~\citep{park2025a} & OGBench cube octuple, state, up to $10^9$ transitions,
      hierarchical & five goals on eight cubes, three rearrangements and two stacks & 8
      $\to$ 8 & 0.36 at 8 over the five goals with $10^8$ transitions and 0.21 with $10^9$;
      0.00 and 0.01 on the two stacking goals with $10^8$, 0.00 on both with $10^9$ \\
    DQC~\citep{li2025dqc} & same benchmark and data as SHARSA, flat action chunking & as
      SHARSA & 8 $\to$ 8 & 0.34 at 8 over the goal
      types, no subtask breakdown \\
    \citet{lampe2024} & ten real Sawyer arms, three workspace cameras with force and torque sensing &
      stacking diverse shapes & 3 in scene, a two object stack, fixed & 0.96 at the fixed
      count \\
    \bottomrule
  \end{tabular}
  \end{center}
\end{table}

\begin{table}[h!]
  \caption{The highest object counts we know of for methods that relax at least one condition
  of Table~\ref{tab:published-learned}, with the same columns. All values are taken from their
  reports.}
  \label{tab:published-relaxed}
  \begin{center}
  \fontsize{7.5}{9}\selectfont
  \renewcommand{\arraystretch}{1.15}
  \begin{tabular}{@{}>{\raggedright\arraybackslash}p{1.9cm}>{\raggedright\arraybackslash}p{3.1cm}>{\raggedright\arraybackslash}p{2.0cm}>{\raggedright\arraybackslash}p{1.9cm}>{\raggedright\arraybackslash}p{3.4cm}@{}}
    \toprule
    \textit{Method} & \textit{Trains on} & \textit{Task} &
    \textit{Counts (train $\to$ test)} & \textit{Best reported} \\
    \midrule
    \citet{janisch2020} & symbolic block world, relational graph, no physics
      simulation & goal-conditioned reconfiguration & 5 $\to$ 2--30 & 0.78 at 20 and 0.54 at
      30 \\
    IL-GNN~\citep{lin2021} & PyBullet and a real Franka, privileged object graph, imitation
      over a predefined pick-and-place primitive & stacking, rearrangement & 3--4 $\to$ 9 with
      data &
      0.75 averaged over counts up to 40, no per count value above 9, manipulation
      supplied by the primitive \\
    Blocks Assemble! \citep{ghasemipour2022} & MuJoCo, privileged state, virtual grippers
      that move blocks without grasping & magnetic block blueprints & 2--16 $\to$ 12--16 held
      out &
      1.00 at 6, 0.99 at 12, 0.91 at 16, success up to a rigid motion of the structure \\
    Transporter~\citep{zeng2020} & PyBullet, RGB-D, scripted two pose pick and place &
      palletizing 18 boxes into a zone & 18 $\to$ 18 & 0.98 at 18 with 1000
      demonstrations, scored as the fraction of
      box volume inside the zone, no per box pose \\
    QT-Opt~\citep{kalashnikov2018} & real arms, RGB, closed-loop grasping & clearing a bin &
      4--10 $\to$ 28 & 0.76 over the first 30 grasps of a 28 object bin and 0.96 with
      replacement, no goal pose \\
    Push Anything~\citep{bui2025push} & real arm, contact implicit MPC, controller not learned &
      pushing to per-object goal poses & 1--4 & 0.98 at 2 and 0.79 at 4 within 2 cm and
      0.1 rad \\
    \bottomrule
  \end{tabular}
  \end{center}
\end{table}

Tables~\ref{tab:published-learned} and~\ref{tab:published-relaxed} list, for every count generalization method cited in
Section~\ref{sec:related} and every gray series of Figure~\ref{fig:results}, what it trains
on, the counts it trains and tests with, and the best result it reports at its highest
evaluated count. A dash marks a quantity the paper does not report in a comparable form.

Every result at a high object count uses privileged state. The only pixel-based method, ECRL,
pushes on a plane without gripper actuation, and its successor HECRL, which grasps and stacks,
completes $0.10$ at six objects from images. Several of the best numbers hold under a criterion less strict than
whole-task success, namely one target cube of six for PAIR and the fraction of goals
filled for IL-GNN. Methods that learn the manipulation itself, in a physics simulator or on a real robot and
with a goal pose for every object, publish whole-task success at up to eight objects. The
offline methods on cube octuple reach that count with 0.21 to 0.36 over the five goals and at most 0.01
on the stacking goals. Every higher count we know of relaxes one of these conditions
(Table~\ref{tab:published-relaxed}). Blocks Assemble! moves blocks with virtual grippers and
scores a blueprint up to a rigid motion. Transporter and IL-GNN execute a scripted primitive
between network-chosen poses and score partial credit. \citet{janisch2020} act in a symbolic
block world without physics simulation. Bin clearing policies such as QT-Opt remove objects
that have no goal pose. Push Anything
reaches four objects with a model-based controller. \citet{li2020} test up to nine blocks, split over several towers, and
show the result as a figure without values, and BuilderBench~\citep{ghugare2025builderbench} reports that its evaluated agents solve two of its twenty-seven hard block structures and that no tabula rasa RL baseline reaches non-zero success.

\section{Ablations and measured constants}
\label{sec:app-ablations}

We ablate WAVE one element at a time on three of the gantry tasks: cube stacking ($h=3$), whose stacks are independent
local structures, the square, whose goals lie flat on the table and constrain each
other only through space, and the pyramid, whose upper goals cannot be filled before the
cubes beneath them are in place. Every ablation keeps the full method and the training budget
of Section~\ref{sec:experiments} and changes exactly one part, either the
selection rule, the side of $\mu$'s interface the warp may edit, or the component of the
warp the policy may use. Each ablation is trained from scratch with three seeds, with one to seven cubes on cube stacking, one to nine on the square and one
to six on the pyramid.

Table~\ref{tab:ablations} shows the results. The WAVE rows repeat
Table~\ref{tab:full-results} for the same task. We summarize each ablation by the mean of its success
over the counts.

\noindent\begin{minipage}{\linewidth}
\begin{center}
  \captionof{table}{Whole task success per object count for the ablations, on cube stacking ($h=3$),
  the square and the pyramid, the mean over three seeds with the minimum and maximum over the seeds in
  parentheses below it. \textsc{Mean} is the mean over the counts beyond the training range, up to thirty cubes. WAVE is the three seed column of
  Table~\ref{tab:full-results} for the same task. Bold marks the highest mean per task and
  count.}
  \label{tab:ablations}
  \fontsize{7}{8.5}\selectfont
  \setlength{\tabcolsep}{3pt}
  \begin{tabular}{lccccccccc}
    \toprule
    $N$ & 3 & 5 & 7 & 10 & 15 & 20 & 25 & 30 & \textsc{MEAN} \\
    \midrule
    \multicolumn{10}{l}{\emph{cube stacking, $h=3$, trained on one to seven}} \\
    WAVE & \textbf{1.00} & 0.95 & \textbf{1.00} & 0.98 & 0.92 & \textbf{0.95} & 0.63 & 0.48 & 0.82 \\
    & (1.00--1.00) & (0.85--1.00) & (1.00--1.00) & (0.95--1.00) & (0.85--1.00) & (0.90--1.00) & (0.45--0.80) & (0.30--0.60) & (0.78--0.86) \\[1pt]
    nearest-first selection & \textbf{1.00} & \textbf{1.00} & 0.97 & 0.97 & \textbf{0.95} & 0.93 & \textbf{0.75} & \textbf{0.60} & \textbf{0.85} \\
    & (1.00--1.00) & (1.00--1.00) & (0.95--1.00) & (0.95--1.00) & (0.95--0.95) & (0.85--1.00) & (0.70--0.80) & (0.45--0.75) & (0.83--0.88) \\[1pt]
    random selection & 0.30 & 0.00 & 0.00 & 0.00 & 0.00 & 0.00 & 0.00 & 0.00 & 0.00 \\
    & (0.20--0.40) & (0.00--0.00) & (0.00--0.00) & (0.00--0.00) & (0.00--0.00) & (0.00--0.00) & (0.00--0.00) & (0.00--0.00) & (0.00--0.00) \\[1pt]
    input channels only & \textbf{1.00} & \textbf{1.00} & \textbf{1.00} & \textbf{1.00} & 0.93 & 0.85 & 0.70 & 0.42 & 0.81 \\
    & (1.00--1.00) & (1.00--1.00) & (1.00--1.00) & (1.00--1.00) & (0.90--0.95) & (0.85--0.85) & (0.65--0.75) & (0.20--0.65) & (0.77--0.85) \\[1pt]
    output channels only & \textbf{1.00} & 0.97 & 0.82 & 0.55 & 0.23 & 0.12 & 0.00 & 0.00 & 0.22 \\
    & (1.00--1.00) & (0.95--1.00) & (0.80--0.85) & (0.55--0.55) & (0.20--0.25) & (0.10--0.15) & (0.00--0.00) & (0.00--0.00) & (0.19--0.26) \\[1pt]
    multipliers only & 0.97 & 0.97 & \textbf{1.00} & 0.90 & 0.82 & 0.78 & 0.47 & 0.25 & 0.65 \\
    & (0.95--1.00) & (0.95--1.00) & (1.00--1.00) & (0.90--0.90) & (0.75--0.90) & (0.65--0.90) & (0.40--0.55) & (0.15--0.35) & (0.58--0.71) \\[1pt]
    offsets only & 0.97 & \textbf{1.00} & 0.93 & 0.88 & 0.62 & 0.60 & 0.23 & 0.10 & 0.52 \\
    & (0.95--1.00) & (1.00--1.00) & (0.90--0.95) & (0.85--0.90) & (0.55--0.70) & (0.55--0.65) & (0.15--0.30) & (0.05--0.15) & (0.50--0.54) \\[1pt]
    \midrule
    \multicolumn{10}{l}{\emph{square, trained on one to nine}} \\
    WAVE & \textbf{1.00} & \textbf{1.00} & 0.98 & \textbf{0.97} & 0.97 & 0.85 & 0.77 & 0.55 & 0.87 \\
    & (1.00--1.00) & (1.00--1.00) & (0.95--1.00) & (0.95--1.00) & (0.90--1.00) & (0.65--0.95) & (0.65--0.85) & (0.45--0.60) & (0.77--0.92) \\[1pt]
    nearest-first selection & \textbf{1.00} & \textbf{1.00} & \textbf{1.00} & \textbf{0.97} & \textbf{1.00} & \textbf{0.93} & \textbf{0.93} & \textbf{0.95} & \textbf{0.97} \\
    & (1.00--1.00) & (1.00--1.00) & (1.00--1.00) & (0.95--1.00) & (1.00--1.00) & (0.90--0.95) & (0.85--1.00) & (0.90--1.00) & (0.94--0.99) \\[1pt]
    random selection & 0.80 & 0.30 & 0.00 & 0.00 & 0.00 & 0.00 & 0.00 & 0.00 & 0.00 \\
    & (0.80--0.80) & (0.25--0.35) & (0.00--0.00) & (0.00--0.00) & (0.00--0.00) & (0.00--0.00) & (0.00--0.00) & (0.00--0.00) & (0.00--0.00) \\[1pt]
    input channels only & \textbf{1.00} & \textbf{1.00} & 0.97 & \textbf{0.97} & 0.88 & 0.75 & 0.60 & 0.60 & 0.76 \\
    & (1.00--1.00) & (1.00--1.00) & (0.95--1.00) & (0.95--1.00) & (0.85--0.90) & (0.65--0.85) & (0.45--0.75) & (0.60--0.60) & (0.71--0.80) \\[1pt]
    output channels only & \textbf{1.00} & 0.82 & 0.68 & 0.33 & 0.20 & 0.05 & 0.03 & 0.00 & 0.14 \\
    & (1.00--1.00) & (0.65--1.00) & (0.35--1.00) & (0.00--0.65) & (0.00--0.40) & (0.00--0.10) & (0.00--0.05) & (0.00--0.00) & (0.01--0.28) \\[1pt]
    multipliers only & \textbf{1.00} & \textbf{1.00} & 0.95 & 0.95 & 0.95 & 0.75 & 0.78 & 0.60 & 0.80 \\
    & (1.00--1.00) & (1.00--1.00) & (0.95--0.95) & (0.90--1.00) & (0.95--0.95) & (0.70--0.80) & (0.65--0.90) & (0.55--0.65) & (0.77--0.82) \\[1pt]
    offsets only & \textbf{1.00} & 0.97 & 0.93 & 0.78 & 0.72 & 0.45 & 0.20 & 0.17 & 0.44 \\
    & (1.00--1.00) & (0.95--1.00) & (0.90--0.95) & (0.70--0.85) & (0.55--0.90) & (0.30--0.60) & (0.10--0.30) & (0.10--0.25) & (0.34--0.54) \\[1pt]
    \midrule
    \multicolumn{10}{l}{\emph{pyramid, trained on one to six}} \\
    WAVE & \textbf{1.00} & \textbf{1.00} & 0.92 & \textbf{0.90} & 0.60 & 0.35 & 0.20 & \textbf{0.08} & 0.47 \\
    & (1.00--1.00) & (1.00--1.00) & (0.85--0.95) & (0.80--0.95) & (0.45--0.70) & (0.20--0.45) & (0.15--0.25) & (0.05--0.15) & (0.46--0.49) \\[1pt]
    nearest-first selection & \textbf{1.00} & 0.93 & 0.50 & 0.47 & 0.25 & 0.15 & 0.03 & 0.00 & 0.19 \\
    & (1.00--1.00) & (0.90--0.95) & (0.10--0.90) & (0.10--0.85) & (0.00--0.50) & (0.00--0.30) & (0.00--0.05) & (0.00--0.00) & (0.01--0.37) \\[1pt]
    random selection & 0.93 & 0.00 & 0.00 & 0.00 & 0.00 & 0.00 & 0.00 & 0.00 & 0.00 \\
    & (0.90--0.95) & (0.00--0.00) & (0.00--0.00) & (0.00--0.00) & (0.00--0.00) & (0.00--0.00) & (0.00--0.00) & (0.00--0.00) & (0.00--0.00) \\[1pt]
    input channels only & 0.97 & 0.55 & 0.35 & 0.17 & 0.05 & 0.03 & 0.00 & 0.00 & 0.07 \\
    & (0.95--1.00) & (0.30--0.80) & (0.15--0.55) & (0.00--0.35) & (0.00--0.10) & (0.00--0.05) & (0.00--0.00) & (0.00--0.00) & (0.01--0.14) \\[1pt]
    output channels only & 0.95 & 0.75 & 0.57 & 0.42 & 0.20 & 0.05 & 0.00 & 0.00 & 0.15 \\
    & (0.95--0.95) & (0.75--0.75) & (0.50--0.65) & (0.40--0.45) & (0.15--0.25) & (0.00--0.10) & (0.00--0.00) & (0.00--0.00) & (0.11--0.18) \\[1pt]
    multipliers only & 0.97 & \textbf{1.00} & \textbf{0.95} & 0.88 & \textbf{0.72} & \textbf{0.53} & \textbf{0.33} & 0.07 & \textbf{0.54} \\
    & (0.95--1.00) & (1.00--1.00) & (0.90--1.00) & (0.85--0.90) & (0.60--0.85) & (0.40--0.65) & (0.30--0.35) & (0.05--0.10) & (0.52--0.57) \\[1pt]
    offsets only & 0.80 & 0.38 & 0.38 & 0.15 & 0.00 & 0.00 & 0.00 & 0.00 & 0.05 \\
    & (0.65--0.95) & (0.00--0.75) & (0.00--0.75) & (0.00--0.30) & (0.00--0.00) & (0.00--0.00) & (0.00--0.00) & (0.00--0.00) & (0.00--0.11) \\[1pt]
    \bottomrule
  \end{tabular}
\end{center}
\end{minipage}

\begin{figure}[h]
  \centering
  \includegraphics[width=\linewidth]{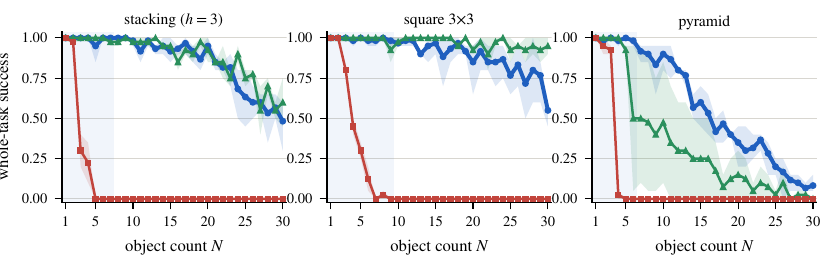}\\[1pt]
  \includegraphics[width=\linewidth]{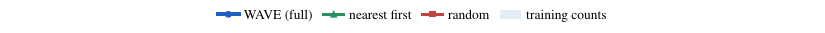}
  \caption{The selection rule, on cube stacking ($h=3$), the square and the pyramid. Random selection
  draws uniformly from the objects whose goal is currently fulfillable. Nearest-first takes the fulfillable object closest to the end effector in the plane. The learned selection policy uses no such filter.
  Shading is the range over the ablation's three seeds, WAVE is its three seed curve, and the
  tinted band marks the task's training counts.}
  \label{fig:abl-selection}
\end{figure}

\runsub{The selection rule}{sec:abl-selection}
Both the random and the nearest-first rule consider only objects whose goal is currently fulfillable, like the scripted selection of the baselines (Appendix~\ref{sec:app-baselines}). The learned selection policy of \eqref{eq:selector} has no such filter. Random selection draws uniformly from these objects. Nearest-first takes the one whose position is closest to the end effector in the plane, an analytic rule that needs no learning and no reward. Figure~\ref{fig:abl-selection} shows the three tasks.

Random selection has the largest effect in the ablations. Its mean success is $0.00$ on all
three tasks, and no seed completes an episode from five cubes on cube stacking, nine on the
square and five on the pyramid.

Nearest-first is close to the learned selection policy. On cube stacking the two lie within
each other's seed range (mean $0.85$ against $0.82$). On the square nearest-first is ahead
($0.97$ against $0.87$) and still reaches $0.95$ at thirty cubes, where WAVE reaches $0.55$.
On the pyramid, whose goals depend on each other, WAVE is ahead ($0.47$ against $0.19$), and
the mean success of the single nearest-first seeds ranges from $0.01$ to $0.37$. Nearest-first only considers objects whose goal is fulfillable, while the learned policy has to find a workable order from
the reward alone. It comes close to the scripted rule and is more reliable on the pyramid.

\begin{figure}[h]
  \centering
  \includegraphics[width=\linewidth]{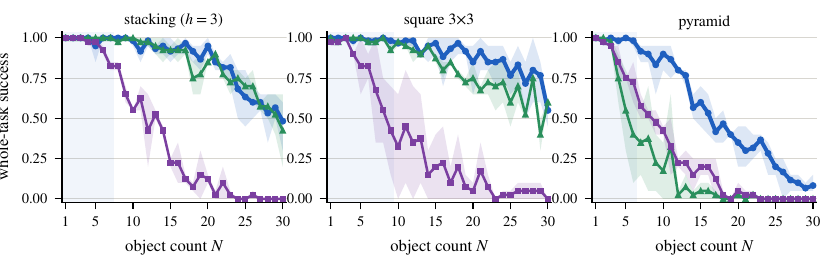}\\[1pt]
  \includegraphics[width=\linewidth]{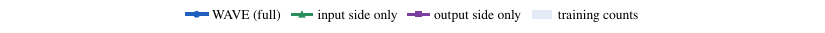}
  \caption{The side of $\mu$'s interface the warp may edit. Input channels only edits what
  $\mu$ observes and executes its action unedited (94 parameters). Output channels only
  edits what $\mu$ commands and shows it the scene as the presentation built it (16). Bands as in
  Figure~\ref{fig:abl-selection}.}
  \label{fig:abl-side}
\end{figure}

\runsub{The warp side}{sec:abl-side}
The warp emits an affine edit of all $55$ channels of $\mu$'s interface, $110$ parameters
in total (Appendix~\ref{sec:app-setup} lists the channels). The input-only ablation restricts it to the $47$ channels $\mu$ reads, so
$\mu$ is shown an edited scene but its action is executed unedited ($94$ parameters). The
output-only ablation restricts it to the $8$ channels $\mu$ commands, so $\mu$ sees the
scene as the presentation built it and only the command stream is edited ($16$ parameters).
Figure~\ref{fig:abl-side} shows the three tasks.

On cube stacking, the input-only warp stays within the seed range of the full warp at nearly
every count (mean $0.81$ against $0.82$). The output-only warp drops from seven cubes
($0.82$), halves by ten ($0.55$) and reaches a mean of $0.22$. The square shows the same
order ($0.76$ input only, $0.14$ output only, $0.87$ full warp). On these two tasks, editing
the input of the skill is close to sufficient. An output side warp can slow, stop or replace
the action of $\mu$, but it cannot change the scene $\mu$ is shown, so $\mu$ still moves
toward whatever stands in its way.

On the pyramid, the input-only warp reaches $0.07$, below the output-only warp at $0.15$,
and both are far below the full warp at $0.47$. Here cubes rest on each other, so the release
itself has to change. A cube has to be set down on its support and let go without disturbing
it, which is an edit of the output. Neither side is sufficient alone on this task, and only
the full warp works on all three tasks.

In Appendix~\ref{sec:app-necessity}, switching off either side of the full warp slows WAVE
down, even on cube stacking. When both sides are trained together, the learned warp uses both. When only one side is trained, that side alone can often solve the task.

\begin{figure}[h]
  \centering
  \includegraphics[width=\linewidth]{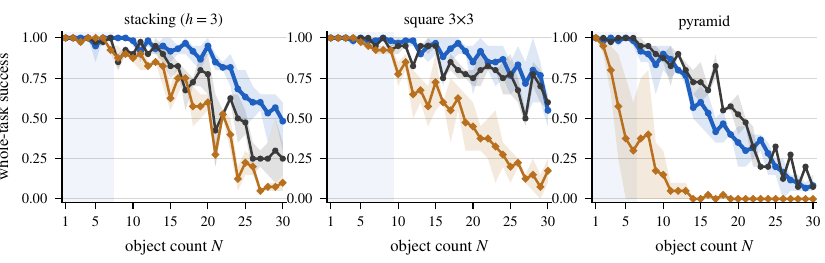}\\[1pt]
  \includegraphics[width=\linewidth]{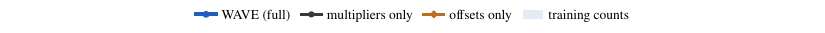}
  \caption{The component of the affine warp the policy may emit. Multipliers only fixes
  every offset at zero and offsets only fixes every multiplier at one, 55 parameters each
  against the full 110. Bands as in Figure~\ref{fig:abl-selection}.}
  \label{fig:abl-component}
\end{figure}

\runsub{The warp component}{sec:abl-component}
The affine warp $z \mapsto m \odot z + o$ has $55$ multipliers and $55$ offsets. The
multipliers-only ablation fixes $o = 0$, and the offsets-only ablation fixes $m = 1$
(Figure~\ref{fig:abl-component}).

Multipliers only reaches a mean success of $0.65$, $0.80$ and $0.54$ on cube stacking, the
square and the pyramid, against $0.82$, $0.87$ and $0.47$ for the full warp. Offsets only
reaches $0.52$, $0.44$ and $0.05$. On the pyramid it already fails within the training counts
($0.38$ at five cubes).

A multiplier can scale or damp a channel, and at $m = 0$ take it over. An offset can shift a
channel to any value, but it cannot make $\mu$ act more slowly or weakly. On the square and
the pyramid the multipliers alone carry the extension. Cube stacking needs both halves.

\runsub{Measured progress per attempt}{sec:abl-constants} We measure the constants of Theorem~1 per attempt on evaluation episodes of the trained policies. An attempt is the period in which one object stays selected (Section~\ref{sec:problem}). $p$ is the fraction of attempts that end with the selected cube placed, $\lambda$ the mean number of previously placed cubes that an attempt knocks off their goals, and $\lmax$ the largest such number. Table~\ref{tab:constants} reports averages over the evaluation episodes. Since Theorem~1 requires bounds that hold after every history, the table serves as a diagnostic and does not verify the assumptions of the theorem. In at most 0.14\% of the attempts, the nearest neighbor of the selected cube is closer than any neighbor seen at the training counts.

\begin{table}[h]
  \caption{Averages per attempt on cube stacking ($h=3$, trained on one to seven cubes) and on the pyramid (trained on one to ten): probability $p$ that the attempt places its cube, number $\lambda$ of placed cubes it knocks off their goals, net progress $\rho = p - \lambda$, largest number $\lmax$ knocked off by a single attempt, and whole-task success of the same episodes. $\rho$ is computed from the unrounded values. The values are averages over episodes and do not verify the assumptions of Theorem~1.}
  \label{tab:constants}
  \begin{center}
  \footnotesize
  \setlength{\tabcolsep}{3pt}
  \begin{tabular}{@{}lccccc@{}}
    \multicolumn{6}{c}{cube stacking, $h=3$} \\
    \toprule
    $N$ & $p$ & $\lambda$ & $\rho$ & $\lmax$ & success \\
    \midrule
    3  & 0.81 & 0.17 & $+0.64$ & 2 & 1.00 \\
    7  & 0.84 & 0.32 & $+0.52$ & 4 & 0.98 \\
    10 & 0.79 & 0.41 & $+0.37$ & 7 & 0.91 \\
    20 & 0.75 & 0.55 & $+0.20$ & 6 & 0.73 \\
    30 & 0.76 & 0.64 & $+0.11$ & 9 & 0.41 \\
    \bottomrule
  \end{tabular}\hfill
  \begin{tabular}{@{}lccccc@{}}
    \multicolumn{6}{c}{pyramid} \\
    \toprule
    $N$ & $p$ & $\lambda$ & $\rho$ & $\lmax$ & success \\
    \midrule
    3  & 0.76 & 0.24 & $+0.51$ & 2 & 0.98 \\
    7  & 0.71 & 0.31 & $+0.39$ & 4 & 1.00 \\
    10 & 0.67 & 0.43 & $+0.25$ & 6 & 0.93 \\
    20 & 0.63 & 0.64 & $-0.01$ & 10 & 0.19 \\
    30 & 0.61 & 0.64 & $-0.03$ & 12 & 0.01 \\
    \bottomrule
  \end{tabular}
  \end{center}
\end{table}

\section{Interventional estimation of warp necessity}
\label{sec:app-necessity}

\begin{figure}[h]
  \centering
  \includegraphics[width=\linewidth]{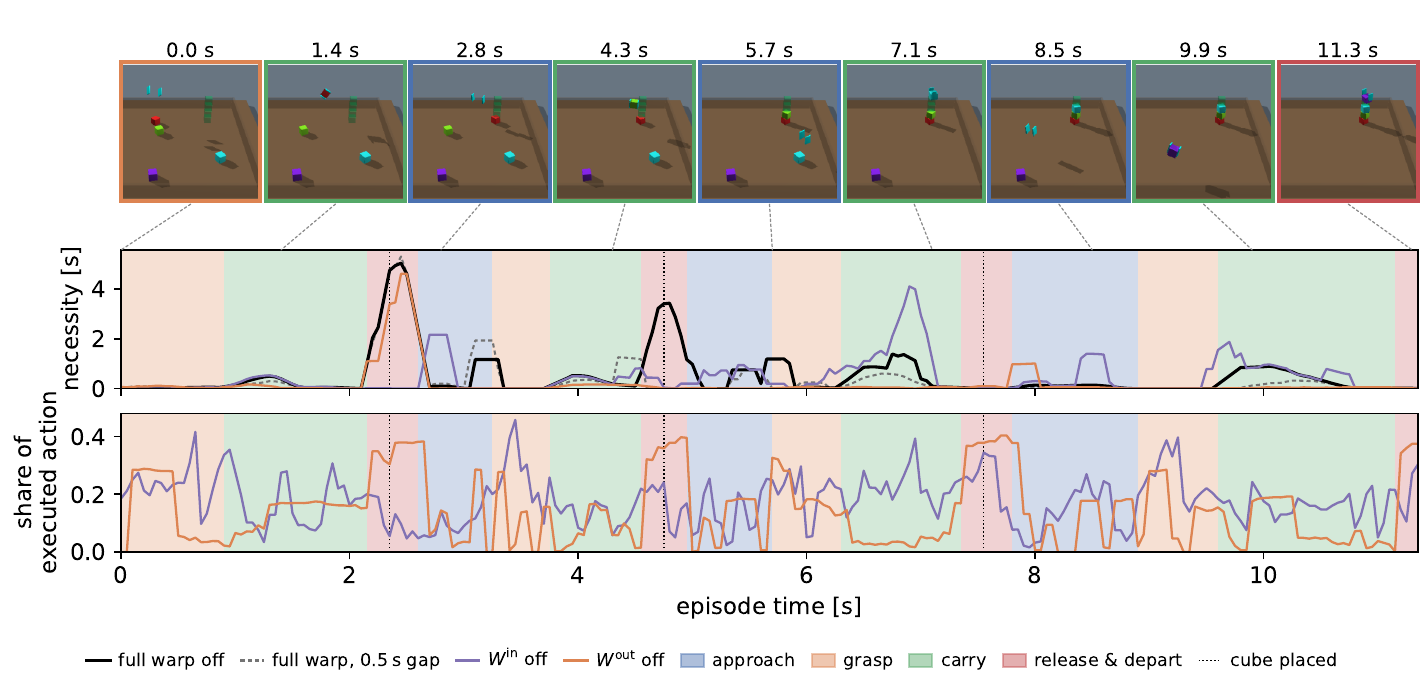}
  \caption{One cube stacking episode with $h=6$, four cubes in one stack. Top, scene views
  framed in the color of the manipulation phase. Middle, necessity, the extra time the current
  phase takes when the full warp (black), only $\Win$ (purple) or only $\Wout$ (orange) is held
  at the identity, or the full warp pauses for $0.5$\,s (dashed gray). Bottom, contribution, the share of the executed action that each side accounts for.}
  \label{fig:necessity}
\end{figure}

To measure when each side of the warp is needed, we split each episode into the phases approach, grasp, carry, release
and depart, read from the simulator state. At every step we copy the simulator and continue the copies
with one of four changes, six copies for each. Either the full warp, only $\Win$ or only $\Wout$ is
held at the identity, or the full warp pauses for $0.5$\,s. The \emph{necessity} of a change is the
extra time the current phase then takes, compared with six unchanged copies. A copy that does
not finish the phase within $6$\,s, or knocks over a placed cube, counts as $6$\,s.

We also measure the \emph{contribution} of each side, how much it changes the executed action.
From the same state we compute the skill action $a_\mu$, the action $a_{\mathrm{in}}$ under
the input warp only, the action $a_{\mathrm{out}}$ under the output warp only, and the executed
action $a$ under both. The two-player Shapley value splits $a - a_\mu$ exactly between the sides,
\begin{equation*}
  \Delta_{\mathrm{in}} = \tfrac{1}{2}\big[(a_{\mathrm{in}} - a_\mu) + (a - a_{\mathrm{out}})\big],
  \qquad
  \Delta_{\mathrm{out}} = \tfrac{1}{2}\big[(a_{\mathrm{out}} - a_\mu) + (a - a_{\mathrm{in}})\big],
\end{equation*}
and the contribution of a side is $\lVert \Delta \rVert$ relative to the diagonal $2\sqrt{d_a}$ of
the action range $[-1, 1]^{d_a}$.

Over $32$ held-out episodes with four cubes, a phase takes about a second. Switching $\Wout$
off costs $1.05$\,s in release and depart, but only $0.27$--$0.34$\,s in the other phases.
Switching $\Win$ off costs $0.95$--$1.56$\,s in every phase.

Figure~\ref{fig:necessity} shows one episode. $\Win$ accounts for $20$--$40\%$ of the executed action throughout, but its necessity stays low during approach and carry, where $\mu$'s own feedback
absorbs a short gap. At the releases the necessity rises, and it comes almost entirely from
$\Wout$. $\mu$ is a grasp and hold policy and never lets go, so $\Wout$ overrides its action to
release the cube (Section~\ref{sec:exp-Ablation}). The output side thus adds a missing behavior at
a few critical moments, and the input side steers $\mu$ continuously.

\section{Formal statements and proofs}
\label{sec:app-proofs}

This appendix answers two questions. First, when does an attempt of the warped skill behave
like an episode of the skill alone, so that it inherits the skill's success
(Lemma~\ref{lem:inherit} and Proposition~\ref{prop:edits})? Second, given how often an attempt
places its object and how many placed objects it knocks off their goals, how many attempts
complete a task of $N$ objects (Theorem~\ref{thm:completion} and
Corollary~\ref{cor:expected})? Neither result predicts how often an attempt of the learned
policy succeeds, which we measure in Table~\ref{tab:constants}.

\paragraph{Setting.} We recall Section~\ref{sec:problem}. The skill $\mu$ is trained in the
single-object MDP $M_1 = (\mathcal{S}_1, \mathcal{A}, P_1, R_1, \mathcal{G})$. We let $P_1$ act
on a (state, goal) pair by carrying the state and leaving the goal fixed. A \emph{solo episode}
is an episode of $\mu$ alone in $M_1$, and $p_1(s_1, g)$ is the probability that a solo episode
started at $(s_1, g)$ brings the object to $g$. $\mathcal{D}_1$ is the set of pairs that the
training of $\mu$ visited, on which $p_1$ is high. In the $N$-object MDP $M_N$, the
presentation $\sigma(s, x^*) \in \mathcal{S}_1 \times \mathcal{G}$ is the scene state $s$ with
every object but the selected $x^*$ deleted, together with the goal of $x^*$. The warp edits
every channel $z$ of the presentation and of $\mu$'s action as $z \leftarrow m z + o$. An
\emph{attempt} is the period in which the selected object stays the same.

\begin{assumption}[the skill]
\label{ass:skill}
$\mu$ is deterministic and has no memory, so its action depends only on the pair it is shown.
\end{assumption}

\begin{assumption}[an undisturbed attempt]
\label{ass:isolated}
During the attempt under consideration no other object touches $x^*$ or the gripper, so the
robot and $x^*$ move as in $M_1$: if the command $a$ is executed in the scene state $s$, the
next presentation is distributed as $P_1(\sigma(s, x^*), a)$.
\end{assumption}

\paragraph{When a warped attempt is an episode of the skill.}

It is not enough that every frame shown to $\mu$ is a valid single-object state. A warp that
shows the object 2\,cm below its true position shows plausible frames at every step, yet no solo
episode produces that sequence, and $\mu$ closes the gripper above the surface. The whole
sequence shown to $\mu$ must be one that $\mu$ could have produced on its own.

\begin{definition}[solo-equivalent edit]
\label{def:solo}
An input edit $\Win$ of (state, goal) pairs and an output edit $\Wout$ of actions are
\emph{solo-equivalent} if editing the next state equals stepping from the edited state with
the unedited action,
$\Win P_1(s_1, \Wout a) = P_1(\Win s_1, a)$ for every pair $s_1$ and action $a$. This is
\eqref{eq:absorb} with $P_1' = P_1$.
\end{definition}

\begin{lemma}[inheritance]
\label{lem:inherit}
Let Assumptions~\ref{ass:skill} and~\ref{ass:isolated} hold for an attempt that starts in the
scene state $s$, and let the warp present $\Win \sigma(s_t, x^*)$ to $\mu$ and execute $\Wout$
of $\mu$'s action at every step $t$ of the attempt, with one solo-equivalent pair
$(\Win, \Wout)$. Then the pairs shown to $\mu$ form a solo episode started at
$\Win \sigma(s, x^*)$. In particular the presented object reaches the presented goal with
probability $p_1(\Win \sigma(s, x^*))$, whatever the number $N$ of objects in the scene.
\end{lemma}

\begin{proof}
Write $y_t = \Win \sigma(s_t, x^*)$ for the pair shown at step $t$. The skill emits
$a_t = \mu(y_t)$ and the robot executes $\Wout a_t$. By Assumption~\ref{ass:isolated} the next
presentation is distributed as $P_1(\sigma(s_t, x^*), \Wout a_t)$, so
$y_{t+1}$ is distributed as $\Win P_1(\sigma(s_t, x^*), \Wout a_t) = P_1(y_t, a_t)$ by
Definition~\ref{def:solo}. This is the transition of a solo episode in which $\mu$ acts on
$y_t$, and by Assumption~\ref{ass:skill} nothing else enters $\mu$'s action.
\end{proof}

\begin{proposition}[what a per-channel affine edit preserves, bounds and replaces]
\label{prop:edits}
Let Assumptions~\ref{ass:skill} and~\ref{ass:isolated} hold for an attempt, and let the warp
keep one edit $z \leftarrow m z + o$ per channel over the attempt.
\begin{enumerate}
  \item[(a)] \emph{Exact edits.} Each of the following is solo-equivalent, so
  Lemma~\ref{lem:inherit} applies.
  \begin{enumerate}
    \item[(a1)] \emph{Retargeting.} An offset on the presented goal channels and the identity
    on all others replaces the goal $g$ by $\tilde{g} = g + o$. The attempt is a solo episode
    toward $\tilde{g}$ and reaches $\tilde{g}$ with probability $p_1(s_1, \tilde{g})$, where $s_1$ is the
    presented state at the start of the attempt.
    \item[(a2)] \emph{Translation.} Let the robot be one whose dynamics do not change under a
    translation parallel to the floor, such as the gantry inside its travel limits. Equal
    offsets on the presented positions of the goal, the object and the gripper translate the
    presented scene. The object reaches its true goal exactly when the presented object
    reaches the presented goal.
    \item[(a3)] \emph{Recalibration.} Let the deployment differ from $M_1$ by a per-channel
    affine recalibration $C_{\mathrm{in}}$ of the sensors and $C_{\mathrm{out}}$ of the
    actuators, $P_1'(s_1, a) = C_{\mathrm{in}}^{-1} P_1(C_{\mathrm{in}} s_1, C_{\mathrm{out}} a)$.
    The constant warp $\Win = C_{\mathrm{in}}$, $\Wout = C_{\mathrm{out}}^{-1}$ satisfies
    \eqref{eq:absorb}, and the conclusion of Lemma~\ref{lem:inherit} holds with $P_1'$ in place
    of $P_1$ in Assumption~\ref{ass:isolated}.
  \end{enumerate}
  \item[(b)] \emph{A bound under an assumption.} Assume that constant biases $b_{\mathrm{in}}$
  on $\mu$'s input and $b_{\mathrm{out}}$ on its output with norms at most $\delta$ lower the
  solo success by at most
  $L_{\mathrm{in}} \|b_{\mathrm{in}}\| + L_{\mathrm{out}} \|b_{\mathrm{out}}\|$. Then an
  attempt under an exact edit of (a) with additional constant offsets $o_{\mathrm{in}}$ and
  $o_{\mathrm{out}}$ of norms at most $\delta$ reaches the presented goal with probability at
  least
  $p_1(\Win \sigma(s, x^*)) - L_{\mathrm{in}} \|o_{\mathrm{in}}\| - L_{\mathrm{out}} \|o_{\mathrm{out}}\|$.
  This is an error bound under the stated assumption and does not follow from
  \eqref{eq:absorb}.
  \item[(c)] \emph{Take-over.} At $m = 0$ on an output channel the executed value is $o$,
  whatever $\mu$ outputs. Take-over is control expressivity only and inherits no guarantee
  from the frozen skill: \eqref{eq:absorb} does not hold, and whether the attempt succeeds is
  learned from the task reward and measured (Table~\ref{tab:ablations},
  Appendix~\ref{sec:app-necessity}).
\end{enumerate}
For any other edit, a multiplier outside $\{0, 1\}$ that is not a recalibration or a bias
larger than $\delta$, we make no statement.
\end{proposition}

\begin{proof}
(a1) $P_1$ neither reads nor changes the goal, so replacing the goal commutes with $P_1$, with
$\Wout$ the identity. (a2) By the assumed invariance, translating the next state equals
stepping from the translated state with the same action, and the translation moves the object
and its goal by the same amount. (a3) Inserting the two maps gives
$C_{\mathrm{in}} P_1'(s_1, C_{\mathrm{out}}^{-1} a)
= C_{\mathrm{in}} C_{\mathrm{in}}^{-1} P_1(C_{\mathrm{in}} s_1, C_{\mathrm{out}} C_{\mathrm{out}}^{-1} a)
= P_1(C_{\mathrm{in}} s_1, a)$, and the proof of Lemma~\ref{lem:inherit} goes through with
$P_1'$ as the dynamics of the attempt. (b) By (a) the attempt is a solo episode with the two
offsets injected as constant biases, and the assumption bounds the loss of success. (c) With
$m = 0$ the action of $\mu$ does not enter the executed value on that channel, so the left
side of \eqref{eq:absorb} does not depend on $a$ there while the right side does.
\end{proof}

\begin{table}[h]
  \caption{The edits of Proposition~\ref{prop:edits}, the behavior each one produces, and what
  follows for the attempt.}
  \label{tab:edits}
  \begin{center}
  \fontsize{7.5}{9}\selectfont
  \setlength{\tabcolsep}{3pt}
  \begin{tabular}{@{}llll@{}}
    \toprule
    edit & channels & behavior & what follows \\
    \midrule
    identity & $m = 1$, $o = 0$ everywhere & $\mu$ unchanged & a solo episode \\
    retargeting (a1) & offset on the goal & detour, hover, retreat, hold & a solo episode to the new goal \\
    translation (a2) & equal offsets on all positions & the same task elsewhere in the arena & $\mu$'s solo success (gantry) \\
    recalibration (a3) & constant $m$, $o$ per channel & other sensor or actuator scaling & $\mu$'s solo success \\
    small constant bias (b) & offset below $\delta$ & slightly shifted grasp or release & a bound, under an assumption \\
    take-over (c) & $m = 0$ on an output channel & release onto a stack, nudging & no statement \\
    any other edit & & & no statement \\
    \bottomrule
  \end{tabular}
  \end{center}
\end{table}

\begin{remark}[re-decided retargets]
\label{rem:retarget}
If the retarget changes during the attempt, the skill starts a new solo episode toward the new
goal from the current state, because it has no memory. At every step $\mu$ then acts on a
state and a goal of $M_1$, which keeps its input of the kind it was trained on. The success
probability of Lemma~\ref{lem:inherit} refers to a retarget that is held until it is reached.
\end{remark}

\begin{remark}[the proposition and the learned policy]
\label{rem:learned}
Proposition~\ref{prop:edits} describes edits that are held over an attempt. The learned warping
policy $\psi$ chooses a new edit at every step and is not restricted to the exact edits, so
the proposition characterizes the class of edits and makes no statement about $\psi$. It
motivates the per-channel affine form, which is the smallest class that contains the identity
(at which training starts), the exact edits of (a) and take-over. Other classes may work as
well or better. Some shifts are outside the class. A rotation of the presented frame about the
vertical requires rotating the action back, which mixes channels, and a change of the size,
shape or dynamics of the object changes $P_1$ itself, so that no edit of $\mu$'s interface maps
it to $M_1$.
\end{remark}

\paragraph{Where the object count can enter.} The learned maps never receive $N$. The warping policy reads the other objects through
per-pair scores reduced by an elementwise minimum, so its input at a fixed local geometry does
not change when objects are added elsewhere, and the selection policy reads a neighborhood of
radius 0.48\,m. The count can then affect the two policies only through the neighborhoods that
occur in a scene of $N$ objects. Whether these remain within what training covered is an
empirical question. Table~\ref{tab:constants} addresses it per count through the success and
the knockdowns of an attempt.

\paragraph{How many attempts complete the task.} Fix the count $N$ and the policy $\pi_N$. Let $\tau_0 = 0 < \tau_1 < \tau_2 < \dots$ be the
control steps at which an attempt ends, because its object is placed or the selection policy
picks another object (Section~\ref{sec:certsel}), so attempt $t$ spans the steps from
$\tau_{t-1}$ to $\tau_t$. An object is \emph{placed} in a scene state when it is within
tolerance of its goal on its intended support. Let $\Phi_t \in \{0, \dots, N\}$ be the number
of placed objects at step $\tau_t$, let $\mathcal{F}_t$ be the history of the episode up to
that step, and let $T_N = \min\{t : \Phi_t = N\}$ be the number of attempts after which the
task is complete. $\Phi_t$ rises by one when an attempt places its object and falls when an
attempt knocks placed objects off their goals.

\begin{assumption}[progress of an attempt]
\label{ass:progress}
There are constants $p_N$, $\lambda_N$ and $\lmax$ such that after every history
$\mathcal{F}_t$ with $\Phi_t < N$ the next attempt ends, places its object with probability at
least $p_N$, places no other object, unplaces at most $\lambda_N$ placed objects in
expectation, and never unplaces more than $\lmax$.
\end{assumption}

The bounds are conditional on the history. They must hold after every history that can occur,
and an average over whole episodes is not enough, since frequent successes early in an episode
could hide later attempts with a smaller expected gain.

\begin{theorem}[conditional completion bound for a fixed $N$]
\label{thm:completion}
Let Assumption~\ref{ass:progress} hold with net progress $\rho_N = p_N - \lambda_N > 0$. Then
for every $c > 1 / \rho_N$ for which $c N$ is an integer,
\begin{equation*}
  \Pr[\,T_N > c N\,] \;\leq\; \exp\!\left( - \frac{2 (\rho_N c - 1)^2 N}{c (1 + \lmax)^2} \right),
\end{equation*}
which is \eqref{eq:bound}.
\end{theorem}

The number of placed objects is a random walk that gains at least $\rho_N$ per attempt in
expectation, so filling $N$ places takes about $N / \rho_N$ attempts, and a budget of $c N$
attempts with $c > 1 / \rho_N$ fails only if the walk falls far behind its expectation.

\begin{proof}
\emph{Step 1, one attempt gains $\rho_N$ objects in expectation.} Fix $t$ with $\Phi_t < N$.
By Assumption~\ref{ass:progress},
$\mathbb{E}[\Phi_{t+1} - \Phi_t \mid \mathcal{F}_t] \geq p_N - \lambda_N = \rho_N$ and
$\Phi_{t+1} - \Phi_t \in [-\lmax, 1]$.

\emph{Step 2, remove the expected gain.} Write $t \wedge T_N = \min(t, T_N)$ and
$M_t = \Phi_{t \wedge T_N} - \rho_N (t \wedge T_N)$. Before $T_N$ the increment of $M_t$ is
$\Phi_{t+1} - \Phi_t - \rho_N$, which is non-negative in expectation by Step~1, and from $T_N$
on it is zero. So $M_t$ is a submartingale, $\mathbb{E}[M_{t+1} \mid \mathcal{F}_t] \geq M_t$,
with increments in $[-\lmax - \rho_N, 1 - \rho_N]$, an interval of length $1 + \lmax$. Azuma's
inequality then gives $\Pr[M_n - M_0 \leq -a] \leq \exp(-2 a^2 / (n (1 + \lmax)^2))$ for every
$a > 0$ and every number of attempts $n$.

\emph{Step 3, failure is such a fall.} If $T_N > c N$, then $\Phi_{cN} < N$ and
$\Phi_0 \geq 0$, hence
$M_{cN} - M_0 = \Phi_{cN} - \Phi_0 - \rho_N c N < -(\rho_N c - 1) N$, which is negative
because $c > 1 / \rho_N$. Step~2 with $n = c N$ and $a = (\rho_N c - 1) N$ gives the bound.
\end{proof}

\begin{corollary}[expected number of attempts]
\label{cor:expected}
Under the hypotheses of Theorem~\ref{thm:completion}, $\mathbb{E}[T_N] \leq N / \rho_N$. The
policy needs at most $1 / \rho_N$ attempts per object in expectation.
\end{corollary}

\begin{proof}
$M_t$ is a submartingale with $M_0 = \Phi_0 \geq 0$, so
$\mathbb{E}[\Phi_{t \wedge T_N}] - \rho_N \mathbb{E}[t \wedge T_N] \geq 0$ for every $t$. Since
$\Phi_{t \wedge T_N} \leq N$, this gives $\mathbb{E}[t \wedge T_N] \leq N / \rho_N$, and
letting $t$ grow gives the claim by monotone convergence.
\end{proof}

\begin{remark}[scope]
\label{rem:scope}
The hypotheses are assumptions on the closed loop of $\pi_N$ at this $N$, and the theorem does
not establish them. Table~\ref{tab:constants} reports averages over evaluation episodes, which
are a diagnostic and do not verify bounds that must hold after every history. The theorem
gives no lower bound on $\rho_N$ that is uniform in $N$, and the measured averages decrease
with $N$. A budget of attempts that is linear in $N$ with one constant for all $N$ would need
such a bound, which we do not have.
\end{remark}

\begin{remark}[how to read the constants]
\label{rem:reading}
The tail bound is loose, because it charges every attempt with the largest knockdown $\lmax$.
With the averages measured at seven cubes on cube stacking ($\rho_N = 0.52$, $\lmax = 4$) and
eight attempts per cube it gives $0.50$, while $0.02$ of the episodes fail. The expected number
of attempts per object, $1 / \rho_N$ (Corollary~\ref{cor:expected}), is more informative. The
theorem counts attempts, whereas an episode is limited in time. An attempt lasts 2.2 to 2.5\,s
on average at every count, so the limit of 20\,s per cube allows about eight attempts per cube,
which requires a net progress of about $0.12$ or more.
\end{remark}

\paragraph{Relation to value-based bounds.} \citet{haramati2024} bound the error of a scene
value that attends over every entity, and this error grows linearly in $N$. Theorem~\ref{thm:completion}
bounds a different quantity, the number of attempts, and contains no approximated scene value.
It depends on $N$ through the constants $p_N$ and $\lambda_N$, which we measure but do not
bound. The idea that a learned quantity defined on a local neighborhood can transfer across
problem sizes has been used before~\citep{ruiz2020,qu2020,yehudai2021}. In WAVE it motivates
the design and is tested empirically.

\section{Positioning of interface-editing methods}
\label{sec:app-positioning}

\begin{table}[h]
  \caption{Interface-editing methods by where they edit a frozen policy, in what form, what the edit is a function of, and which shift it is meant to absorb. Proposition~\ref{prop:edits} states for which edits of WAVE the behavior of the skill is preserved.}
  \label{tab:positioning}
  \begin{center}
  \fontsize{7.5}{9}\selectfont
  \renewcommand{\arraystretch}{1.15}
  \begin{tabular}{@{}>{\raggedright\arraybackslash}p{2.4cm}>{\raggedright\arraybackslash}p{1.8cm}>{\raggedright\arraybackslash}p{2.2cm}>{\raggedright\arraybackslash}p{3.2cm}>{\raggedright\arraybackslash}p{2.4cm}@{}}
    \toprule
    \textit{Method} & \textit{Edit site} & \textit{Edit form} &
    \textit{The edit is a function of} & \textit{Shift the edit absorbs} \\
    \midrule
    Residual RL~\citep{silver2018,johannink2019,yuan2024} & action & additive (bounded in \citeauthor{yuan2024}) & the current state &
      same task, refined \\
    Warp RL~\citep{hirschowitz2026} & action distribution & monotone spline & the current state & dynamics \\
    CLAE~\citep{das2026} & internal activations & affine & the current state & behavior mode \\
    ZPRL~\citep{yu2026} & inserted latent & latent perturbation & the context, once per action chunk & same task, refined \\
    DC-Ada~\citep{dcadand} & observation & residual network found by reward-only random search & the current observation, with parameters fixed per robot &
      sensor heterogeneity \\
    APB~\citep{policynda} & before and after a backbone & linear & nothing; fixed per task &
      task (meta-transfer) \\
    Pivoting~\citep{zhang2023b} & observation and action & diagonal scaling & observation: a descriptor of the manipulated object, constant over an episode. Action: that descriptor and the current state & shape and size of the manipulated object \\
    Inter-task mappings~\citep{taylor2007} & state and action & hand-specified map & nothing; fixed per task pair &
      task pair, including more agents; the value function is trained further \\
    This work & observation and action & per-channel discrete affine & the other objects around the selected one, at every step &
      \textit{scene composition (object count)} \\
    \bottomrule
  \end{tabular}
  \end{center}
\end{table}

Prior work edits the observation of a frozen policy~\citep{dcadand,policynda,zhang2023b}, both
its input and its output~\citep{zhang2023b,policynda,taylor2007}, or its internal activations
with an affine edit chosen by an RL policy~\citep{das2026}. These edits adapt a policy to new sensors, a new task, a new object or a new behavior mode (Table~\ref{tab:positioning}), and none of them changes the number of objects the fixed policy has to handle. The inter-task mappings of \citet{taylor2007} transfer between tasks with different numbers of agents, from three against two to five against four players, but with a hand-specified mapping and further training in the target task. WAVE is closest to \citet{zhang2023b}, from which it takes the frozen skill, per-channel edits of its input and its action that are trained with a policy gradient, and an action edit that depends on the current state. It differs in four points. (i) Both edits are conditioned on the objects around the selected one, whereas the pivoting setting contains a single object and the edits depend on a descriptor of that object. (ii) The input edit changes during an episode. When we fix it per episode, as their state projection is, the policy still learns its training counts, but the mean success beyond the training range drops from $0.82$ to $0.24$ on cube stacking and from $0.96$ to $0.43$ on the line (Section~\ref{sec:exp-baselines}). (iii) Every channel has an offset in addition to the multiplier, which raises the mean success on cube stacking from $0.65$ to $0.82$ and makes no difference on the square and the pyramid (Table~\ref{tab:ablations}). (iv) The selection policy is learned, which improves on a scripted nearest-first rule on the pyramid ($0.47$ against $0.19$) but not on the other two tasks. The ablations of \citet{zhang2023b} rank the two sides of the interface in the same order as ours. Their pivoting skill transfers in 30 of 40 trials with both projections, in 27 without the action projection, in 19 without the state projection, and in none when the diagonal structure is replaced by free-form neural maps.

The classical transfer literature contains related framings that edit no interface. MDP
homomorphisms and inter task transfer~\citep{ravindran2003,taylor2009} relate two tasks
through structural correspondences, options and learned initiation sets~\citep{sutton1999,graves2020a} decide when a fixed sub policy may run, and prior policies
shape the exploration of a new task~\citep{uchendu2022,schmitt2018,galashov2019,tirumala2020}. None of them edits what a skill reads or emits, so Table~\ref{tab:positioning} does not list them.

\section{Generalization experiments}
\label{sec:app-generalization}

Section~\ref{sec:experiments} varies the number of cubes and keeps everything else fixed. The
eight experiments of this section each change one of these fixed parts. Every experiment has one
paragraph, which states in this order the question, how the experiment differs from
Section~\ref{sec:experiments}, the result, and what the result shows. Unless a paragraph says
otherwise, the setting is that of Section~\ref{sec:exp-count}: the gantry, the frozen skill $\mu$
trained on a single cube with an edge of 16\,cm, whole-task success in the first episode of each
scene with deterministic policies, and a time limit of 20\,s per object. Only the warping policy
and the selection policy are trained. Where a paragraph says that the training is
\emph{continued}, the two policies of the first paragraph are trained for a further
$4.0 \times 10^8$ environment steps on scenes with 1 to 10 objects and evaluated in 64 scenes per
count.

\begin{figure}[!htb]
  \centering
  \includegraphics[width=0.8\linewidth]{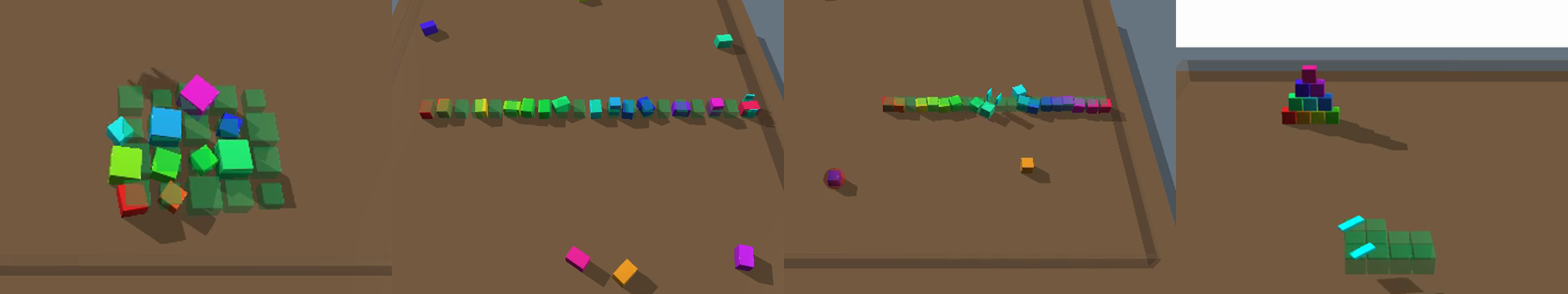}
  \caption{From left to right: cubes of three sizes (square), boxes (line), a line episode in
  which a disturbance has set a cube to another position (red marker), and a built pyramid at the
  start of an episode whose goal is the square.}
  \label{fig:generalization-renderings}
\end{figure}

\begin{figure}[!htb]
  \centering
  \includegraphics[width=\linewidth]{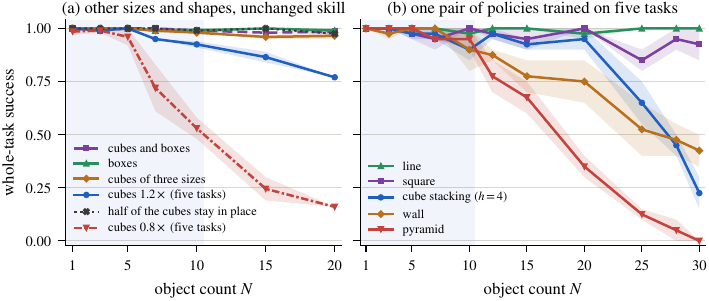}
  \caption{Whole-task success against the object count $N$. The blue
  band marks the training counts. (a) Continued training with objects of other sizes and shapes
  and with cubes that have to stay in place. (b) One warping policy and one
  selection policy trained on five tasks at once, 20 scenes per count.}
  \label{fig:generalization}
\end{figure}

\paragraph{One pair of policies for five tasks.} Section~\ref{sec:experiments} trains a separate
warping policy and selection policy for every task, and we ask whether one pair can learn several
tasks. We train one warping policy and one selection policy on the line, the square, the wall,
the pyramid and cube stacking ($h=4$) at once. Every training scene shows one of the five tasks
with 1 to 10 cubes, and the training takes $1.57 \times 10^9$ environment steps, compared to
$5.0 \times 10^8$ for one task in Section~\ref{sec:experiments}. The inputs of both policies are
unchanged, so they can tell the tasks apart only by the goal poses. This pair completes at least
$0.90$ of the episodes of every task with up to ten cubes. Beyond the training counts it completes
$1.00$ of the line episodes and $0.93$ of the square episodes at thirty cubes and $0.95$ of the
stacking episodes at twenty, while the wall and the pyramid fall to $0.75$ and $0.35$ at twenty
(Figure~\ref{fig:generalization}b, Table~\ref{tab:generalization}). Training one pair of policies
per task is therefore not a requirement of WAVE. The numbers do not show whether training on
five tasks is better or worse than training on one, because the training counts, the stack
height and the workspace, which is larger for the square, the wall and the pyramid, differ from
those of Table~\ref{tab:full-results}.

\begin{table}[!htb]
  \caption{Whole-task success in five of the generalization experiments.  The tasks of the training are
  in parentheses in the row labels. Disturbances: policies
  of Table~\ref{tab:full-results} evaluated with a time limit of
  40\,s per cube. All values are the mean of three seeds.}
  \label{tab:generalization}
  \centering
  \fontsize{7.5}{9}\selectfont
  \begin{tabular}{lcccc}
    \toprule
    objects of other sizes and shapes & $N=5$ & $N=10$ & $N=15$ & $N=20$ \\
    \midrule
    cubes with 1.2 times the edge (five tasks)                 & 1.00 & 0.93 & 0.87 & 0.77 \\
    cubes with 0.8 times the edge (five tasks)                 & 0.96 & 0.53 & 0.25 & 0.16 \\
    boxes of $16 \times 24 \times 16$\,cm (line, square)       & 1.00 & 0.99 & 1.00 & 0.99 \\
    cubes of 0.8, 1.0 and 1.2 times the edge in one scene (line, square) & 1.00 & 0.98 & 0.96 & 0.97 \\
    cubes and boxes in one scene (line, square)                & 1.00 & 0.99 & 0.98 & 0.99 \\
    \midrule
    cubes that have to stay in place & $N=5$ & $N=10$ & $N=15$ & $N=20$ \\
    \midrule
    half of the cubes have their start pose as the goal (line, square) & 1.00 & 0.99 & 1.00 & 0.98 \\
    \midrule
    starting from a built structure & $N=3$ & $N=6$ & $N=10$ & $N=15$ \\
    \midrule
    pyramid to square                                          & 0.98 & 0.87 & 0.59 & 0.30 \\
    stacks of height three to line                             & 0.98 & 0.95 & 0.90 & 0.90 \\
    pyramid to square, failure when a cube falls & 0.92 & 0.42 & 0.10 & 0.00 \\
    stacks of height three to line, failure when a cube falls & 0.82 & 0.64 & 0.56 & 0.40 \\
    \bottomrule
  \end{tabular}

  \vspace{0.4em}
  \begin{tabular}{lcccccc}
    \toprule
    & \multicolumn{3}{c}{$N=10$} & \multicolumn{3}{c}{$N=20$} \\
    \cmidrule(lr){2-4} \cmidrule(lr){5-7}
    one disturbance every & never & 20 s & 10 s & never & 20 s & 10 s \\
    \midrule
    cube stacking ($h=3$) & 0.95 & 0.99 & 0.98 & 0.91 & 0.93 & 0.90 \\
    line                  & 0.99 & 0.98 & 0.96 & 0.98 & 0.97 & 0.88 \\
    square                & 0.99 & 0.97 & 0.97 & 0.95 & 0.96 & 0.90 \\
    pyramid               & 0.94 & 0.99 & 0.98 & 0.59 & 0.69 & 0.52 \\
    \midrule
    one pair of policies for five tasks & $N=5$ & $N=10$ & $N=15$ & $N=20$ & $N=25$ & $N=30$ \\
    \midrule
    line                  & 1.00 & 0.98 & 1.00 & 0.98 & 1.00 & 1.00 \\
    square                & 0.98 & 1.00 & 0.95 & 1.00 & 0.85 & 0.93 \\
    cube stacking ($h=4$) & 0.98 & 0.90 & 0.93 & 0.95 & 0.65 & 0.23 \\
    wall                  & 1.00 & 0.90 & 0.78 & 0.75 & 0.53 & 0.43 \\
    pyramid               & 1.00 & 0.95 & 0.68 & 0.35 & 0.13 & 0.00 \\
    \bottomrule
  \end{tabular}
\end{table}

\paragraph{Cubes of another size.} The skill was trained on a cube with an edge of 16\,cm, and
neither the skill nor the two policies receive the dimensions of an object, so we ask whether
WAVE can use this skill for cubes of another size. The training is continued on the five tasks
with every cube scaled to 1.2 times the edge (19.2\,cm) or to 0.8 times the edge (12.8\,cm). The
distance between neighboring goals scales with the cubes, while the gripper and the position
tolerance keep their size. With the larger cubes, whole-task success over the five tasks is
$0.77$ / $0.77$ at twenty cubes, twice the largest training count
(Figure~\ref{fig:generalization}a). The line and the square stay at $0.94$ or above there, and
the wall ($0.67$ / $0.72$) and the pyramid ($0.48$ / $0.36$) account for the loss. With the
smaller cubes, success is $0.92$ / $1.00$ at five cubes, $0.48$ / $0.58$ at ten and $0.16$ /
$0.16$ at twenty. The warp thus adapts the skill to a larger cube but not to the smaller one.
Because the decrease with the smaller cubes begins within the training counts, it is not a
failure to generalize over the count. The failed episodes end at the time limit, with about 100
placed-to-unplaced transitions per episode at twenty cubes. We attribute this to the unchanged
gripper, which has to work between goals that lie closer together, but we did not isolate the
cause.

\paragraph{Boxes, and objects that differ within a scene.} We ask whether the skill for one cube
is also sufficient for an object of another shape and for scenes whose objects are not all the
same. The training is continued on the line and the square with three sets of objects: boxes of
$16 \times 24 \times 16$\,cm, cubes with 0.8, 1.0 and 1.2 times the edge in one scene, and cubes
and boxes in one scene. All three sets reach a whole-task success of at least $0.95$ at twenty
objects, twice the largest training count (Figure~\ref{fig:generalization}a,
Table~\ref{tab:generalization}). A skill trained on one cube is therefore sufficient for boxes
that are 1.5 times as long as the cube and for scenes with objects of different sizes and
shapes. The scenes with three sizes also contain the small cubes of the previous paragraph, so
the skill can place a small cube, and the failure there concerns scenes that consist only of
small cubes. Two limits remain. On the line and the square every object lies on the floor, so
these results do not cover stacking such objects. Objects of other classes, for example
cylinders, were not tested, and a class that the skill cannot manipulate alone would need its
own skill.

\paragraph{Cubes that have to stay in place.} In Section~\ref{sec:experiments} every cube of a
scene has a goal in the structure. We ask whether WAVE can build a structure between cubes that
are not part of it. On the line and the square, half of the $N$ cubes have their own start pose
as the goal, and they start next to the goal positions of the structure. The training is
continued on such scenes. Whole-task success is defined as in Section~\ref{sec:experiments}, so
it requires each of these cubes to be within the tolerance of its start pose when the episode
ends, but it does not forbid contact with them during the episode. Success is $1.00$ / $0.98$ at
ten cubes and $0.98$ / $0.97$ at twenty (Table~\ref{tab:generalization}). The policies thus
complete the structure while the other cubes end at their start poses, also at twice the largest
training count. They do not avoid every contact: at twenty cubes a placed cube is knocked off its
goal and placed again about 33 times per episode, counted over both kinds of cubes.

\paragraph{Disturbances during the episode.} In Section~\ref{sec:experiments} only the robot
changes the scene. We ask whether the policies also complete a task when the scene is changed
from outside while they act. We evaluate the policies of Table~\ref{tab:full-results} without
retraining. Every 20\,s or every 10\,s, a disturbance sets
one cube instantly to a free position on the floor. This cube is at rest, is not in the gripper
and carries no other cube, and it is drawn with equal probability from the placed cubes or from
the cubes not yet placed. Since every disturbance adds work, all episodes of this experiment,
also those without disturbances, have a time limit of 40\,s per cube. Whole-task success is at
most $0.10$ lower than without disturbances (Table~\ref{tab:generalization}), also on the line
with twenty cubes and one disturbance every 10\,s ($0.88$ against $0.98$), where about 45
disturbances occur per episode. The policies handle a disturbance because the warp is computed
from the current crop at every control step and the selection policy decides again whenever a
cube is placed or knocked off its goal (Section~\ref{sec:problem}), so a cube set to another
position is treated like any cube that is not at its goal. The cost is time: with the 20\,s per
cube of Section~\ref{sec:experiments}, success on the line with twenty cubes is $0.98$ without
disturbances, $0.84$ with one every 20\,s and $0.47$ with one every 10\,s. We did not test
objects that move continuously on their own.

\paragraph{Starting from a built structure.} In Section~\ref{sec:experiments} no episode starts
from a built structure. We ask whether WAVE can turn one structure into another, which requires
taking cubes out of a structure. All $N$ cubes start as a built pyramid with the
square as the goal, or as stacks of height three with the line as the goal, and the training is
continued on each of these two tasks. The policies reach the goal structure, in $0.84$ / $0.89$
of the episodes from a pyramid of six cubes and in $0.88$ / $0.92$ from stacks with ten cubes
(Table~\ref{tab:generalization}), but they do so by knocking the start structure down and
collecting the cubes from the floor. To test whether a structure can be taken apart cube by cube,
we train for a further $8.0 \times 10^8$ steps with one added condition: an episode ends as a
failure when a cube that the gripper has not yet held drops more than 4\,cm below its resting
height in the start structure. Under this condition a pyramid is turned into a square in
$0.86$ / $0.98$ of the episodes with three cubes, $0.34$ / $0.50$ with six and $0.02$ / $0.17$
with ten, and stacks into a line in $0.77$ / $0.34$ with ten cubes. WAVE thus reaches a new
structure from a built one, but it learns to take a structure apart without a falling cube
reliably only for three cubes. The skill was trained to pick up a cube that stands free, which
likely explains this limit, but we did not test this explanation.

\label{app:held_scores}
\paragraph{Success that lasts, and success after a release.} In Section~\ref{sec:experiments} an
episode ends as a success at the first step at which every cube is placed, so cubes can still be
in motion and the last cube can still be in the gripper. We ask whether the finished structure
lasts. We evaluate the policies of Table~\ref{tab:full-results} without retraining under two
stricter criteria (three seeds, 64 episodes per seed and count). Under the first, the policies
keep acting after the first success, and an episode counts as a success only when all cubes have
stayed placed for 2\,s without interruption. Its time limit is 50\,s per cube, because a
structure that does not last has to be completed again. Under the second, with the time limit
of Section~\ref{sec:experiments}, a scripted routine opens and lifts the gripper as soon as all
cubes are placed and slower than 0.05\,m/s, and we check 2\,s later whether every cube is still
placed. On cube stacking ($h=3$), success is held for 2\,s in $0.99$ of the episodes at seven
cubes, $0.92$ at twenty and $0.59$ at thirty, compared to $0.99$, $0.94$ and $0.65$ for the
first success within the same time limit (Table~\ref{tab:release}). After the release, all
cubes are placed in $0.93$, $0.73$ and $0.38$ of the episodes at these counts, which is $0.95$,
$0.90$ and $0.85$ of the episodes that reach whole-task success. Although the policies were
trained for neither criterion, the whole-task success of Section~\ref{sec:experiments} thus
mostly corresponds to structures that last. The exceptions are stacks of height six, which
remain after the release in $0.82$ of the successful episodes at seven cubes and in $0.48$ at
fifteen, and the wall, where success is held for 2\,s in $0.82$ of the episodes at five cubes
compared to $0.98$ for the first success. The release loses about $0.1$ already with a single
cube, so part of the loss after the release is caused by the scripted routine and not by the
structure.

\begin{table}[!htb]
  \caption{The policies of Table~\ref{tab:full-results} under two stricter success criteria, mean
  of three seeds with 64 episodes each. Top: whole-task success / fraction of episodes in which
  every cube is still placed 2\,s after a scripted release, with a time limit of 20\,s per cube
  (40\,s for stacks of height six). Bottom: first success / success held for 2\,s without
  interruption, both with a time limit of 50\,s per cube. Stacks of height six were not
  evaluated at 25 and 30 cubes.}
  \label{tab:release}
  \centering
  \fontsize{7.5}{9}\selectfont
  \setlength{\tabcolsep}{4pt}
  \begin{tabular}{@{}lccccc@{}}
    \toprule
    whole-task success / placed after release & $N=1$ & $N=7$ & $N=15$ & $N=20$ & $N=30$ \\
    \midrule
    cube stacking ($h=3$) & 1.00 / 0.89 & 0.97 / 0.93 & 0.90 / 0.85 & 0.81 / 0.73 & 0.45 / 0.38 \\
    cube stacking ($h=6$) & 0.99 / 0.90 & 0.99 / 0.81 & 0.86 / 0.41 & 0.13 / 0.02 & 0.00 / 0.00 \\
    line                  & 1.00 / 0.89 & 1.00 / 0.88 & 0.97 / 0.88 & 0.96 / 0.86 & 0.92 / 0.83 \\
    square                & 0.99 / 0.83 & 0.98 / 0.84 & 0.95 / 0.79 & 0.88 / 0.66 & 0.66 / 0.52 \\
    pyramid               & 0.99 / 0.87 & 0.91 / 0.78 & 0.61 / 0.53 & 0.43 / 0.41 & 0.06 / 0.05 \\
    wall                  & 1.00 / 0.88 & 0.98 / 0.82 & 0.92 / 0.79 & 0.77 / 0.66 & 0.40 / 0.36 \\
    \bottomrule
  \end{tabular}

  \vspace{0.4em}
  \begin{tabular}{@{}lcccccc@{}}
    \toprule
    first success / held for 2\,s & $N=5$ & $N=10$ & $N=15$ & $N=20$ & $N=25$ & $N=30$ \\
    \midrule
    cube stacking ($h=3$) & 0.98 / 0.98 & 0.99 / 0.99 & 0.93 / 0.93 & 0.94 / 0.92 & 0.82 / 0.76 & 0.65 / 0.59 \\
    cube stacking ($h=6$) & 0.99 / 0.99 & 0.97 / 0.97 & 0.88 / 0.68 & 0.17 / 0.05 & -- & -- \\
    line                  & 1.00 / 0.98 & 1.00 / 0.98 & 0.98 / 0.98 & 1.00 / 0.99 & 0.98 / 0.98 & 0.97 / 0.97 \\
    square                & 1.00 / 0.97 & 0.99 / 0.99 & 0.98 / 0.95 & 0.98 / 0.96 & 0.91 / 0.86 & 0.92 / 0.89 \\
    pyramid               & 0.98 / 0.90 & 0.93 / 0.87 & 0.78 / 0.72 & 0.69 / 0.60 & 0.44 / 0.35 & 0.19 / 0.15 \\
    wall                  & 0.98 / 0.82 & 0.98 / 0.82 & 0.97 / 0.83 & 0.91 / 0.77 & 0.87 / 0.69 & 0.61 / 0.53 \\
    \bottomrule
  \end{tabular}
\end{table}

\paragraph{Minimum, mean or attention over the other objects.} The warping policy reads the other
objects through per-pair scores, which an elementwise minimum reduces to one encoding, and
Section~\ref{sec:warper} motivates the minimum by its independence of the count. We ask whether
the count generalization of Section~\ref{sec:experiments} depends on the minimum. We retrain WAVE
on cube stacking ($h=3$) and on the pyramid as in Section~\ref{sec:experiments}, and replace only the minimum. The first replacement is the mean of the scores over the
other objects. The second is an attention-weighted mean, in which a second network maps each
pair to one logit per score, a softmax over the other objects turns the logits into weights, and
the scores are averaged with these weights. On cube stacking, the mean success beyond the
training counts is $0.55$ with the mean and $0.64$ with the attention, compared to $0.82$ with
the minimum (Table~\ref{tab:reduction}). On the pyramid it is $0.47$ with the mean, $0.66$ with
the attention and $0.47$ with the minimum. The minimum is thus the best of the three on cube
stacking but not on the pyramid, so count generalization does not depend on the minimum, and we
do not claim that the minimum is the best choice for every task. All three variants score each of
the other objects separately and give the warping policy an input whose size does not depend on
$N$. We did not compare against self-attention among the other objects or against inputs whose
size grows with $N$.

\begin{table}[!htb]
  \caption{Whole-task success when the elementwise minimum over the other objects is replaced by a
  mean or by an attention-weighted mean. The minimum rows are the policies of
  Table~\ref{tab:full-results}, all rows use three seeds. \textsc{Mean} is the mean over the counts beyond the training counts.}
  \label{tab:reduction}
  \centering
  \fontsize{7.5}{9}\selectfont
  \setlength{\tabcolsep}{4pt}
  \begin{tabular}{@{}lccccccc@{}}
    \toprule
    & $N=7$ & $N=10$ & $N=15$ & $N=20$ & $N=25$ & $N=30$ & \textsc{Mean} \\
    \midrule
    \multicolumn{8}{@{}l}{\emph{cube stacking ($h=3$), trained on one to seven cubes}} \\
    minimum (WAVE) & 1.00 & 0.98 & 0.92 & 0.95 & 0.63 & 0.48 & 0.82 \\
    mean           & 0.95 & 0.88 & 0.80 & 0.35 & 0.40 & 0.05 & 0.55 \\
    attention      & 1.00 & 0.90 & 0.90 & 0.78 & 0.30 & 0.08 & 0.64 \\
    \midrule
    \multicolumn{8}{@{}l}{\emph{pyramid, trained on one to six cubes}} \\
    minimum (WAVE) & 0.92 & 0.90 & 0.60 & 0.35 & 0.20 & 0.08 & 0.47 \\
    mean           & 1.00 & 0.88 & 0.55 & 0.38 & 0.13 & 0.05 & 0.47 \\
    attention      & 1.00 & 0.93 & 0.83 & 0.63 & 0.43 & 0.43 & 0.66 \\
    \bottomrule
  \end{tabular}
\end{table}

\end{document}